\documentclass[11pt]{article}

\usepackage[final]{acl}

\usepackage{times}
\usepackage{latexsym}

\usepackage[T1]{fontenc}
\usepackage[utf8]{inputenc}

\usepackage{microtype}

\usepackage{inconsolata}

\usepackage{graphicx}

\usepackage{bm}
\usepackage{graphicx}
\usepackage{makecell}
\usepackage{arydshln}
\usepackage{amssymb}
\usepackage{subcaption}
\usepackage{amsmath}
\usepackage{tcolorbox}

\DeclareMathOperator*{\argmax}{arg\,max}
\title{From Interpretability to Performance:\\ Optimizing Retrieval Heads for Long-Context Language Models}

\author{Youmi Ma \qquad
  Naoaki Okazaki \\
  Department of Computer Science, Institute of Science Tokyo \\
  \texttt{\{ma.y, okazaki\}@comp.isct.ac.jp} \\}

\begin{document}
\maketitle
\begin{abstract}
Advances in mechanistic interpretability have identified special attention heads, known as retrieval heads, that are responsible for retrieving information from the context.
However, the role of these retrieval heads in improving model performance remains unexplored.
This work investigates whether retrieval heads can be leveraged to enhance the long-context capabilities of LLMs.
Specifically, we propose RetMask, a method that generates training signals by contrasting normal model outputs with those from an ablated variant in which the retrieval heads are masked.
This mechanism-based approach achieves substantial improvements: +2.28 points on HELMET at 128K for Llama-3.1, with +70\% gains on generation with citation and +32\% on passage re-ranking, while preserving performance on general tasks. 
Experiments across four models in three families demonstrate that RetMask consistently improves long-context performance, where gains correlate with the sparsity of the retrieval score distribution: models with sparser distributions, where retrieval capabilities are concentrated in a small set of heads, respond more strongly, while those with less sparse distributions show more modest gains. 
These results validate the functional role of retrieval heads and show that mechanistic insights can be transformed into performance enhancements\footnote{The source code is available at: \url{https://github.com/YoumiMa/RetMask}.}.
\end{abstract}

\section{Introduction}

\begin{figure}[t]
  \includegraphics[width=\columnwidth]{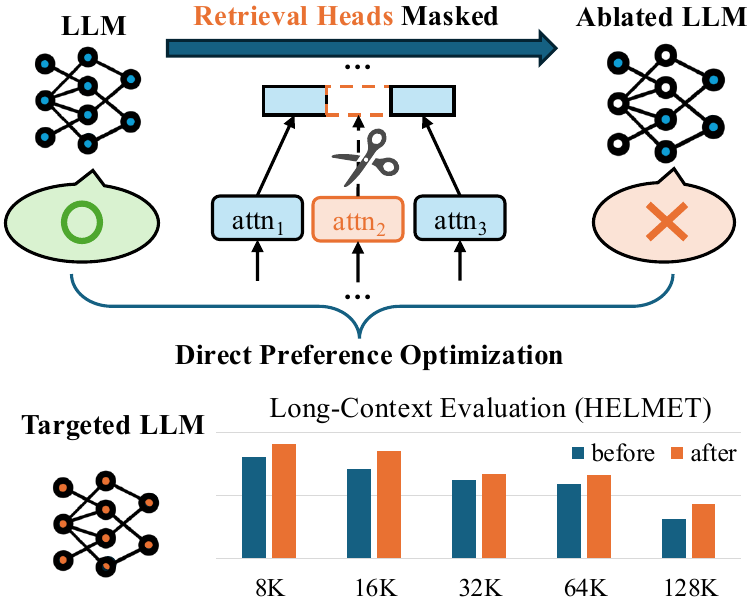}
  \caption{An overview of RetMask. Given an LLM, we construct an ablated variant by masking its retrieval heads~\cite{wu2025retrieval}, then sample responses from both the original and ablated models to construct training pairs for Direct Preference Optimization. We validate this approach on four models across three families. }
  \label{fig:overview}
\end{figure}

Large Language Models (LLMs) require long-context capabilities to realize multi-document understanding~\cite{bai-etal-2024-longbench}, in-context learning~\cite{brown2020languagemodelsfewshotlearners}, and test-time scaling~\cite{snell2024scalingllmtesttimecompute,openaireasoning}.
Recent studies on mechanistic interpretability revealed that long-context factuality is closely related to a set of attention heads named \textit{retrieval heads}~\cite{wu2025retrieval,zhang-etal-2025-query-focused}.
Retrieval heads attend to previous tokens and recall information during the generation process.
Deactivating retrieval heads has been reported to result in performance drops for downstream tasks.

While retrieval heads provide hints for the mechanism of long-context capabilities, their contributions to model performance remain unexplored.
This gap between interpretability and model performance is pervasive: despite identifying specialized components responsible for knowledge storage~\cite{dai-etal-2022-knowledge,meng2022locating} and language~\cite{tang-etal-2024-language,kojima-etal-2024-multilingual}, prior work has not established effective methods to transform these discoveries into performance enhancements.
\citet{gu-etal-2024-model} reports that editing knowledge-specific components brings unintended side effects on models' general abilities, and \citet{mondal-etal-2025-language} reports that language-specific neuron interventions are insufficient to provide performance gains on downstream tasks.
This leads to a natural question: Can retrieval heads be leveraged to enhance long-context capabilities?

With this research question in mind, this paper explores a method to enhance long-context processing abilities by optimizing retrieval heads.
Specifically, as shown in Figure~\ref{fig:overview}, we synthesize supervision data from both the original model and its ablated variant in which the retrieval heads are masked.
We name the method as RetMask, short for \textbf{Ret}rieval-Head \textbf{Mask}ing.
RetMask applies Direct Preference Optimization (DPO,~\citealp{rafailov2023direct}) to post-trained models to prefer responses generated by the original model over those generated by its retrieval-head-ablated counterpart.
Evaluations on HELMET~\cite{yen2025helmet} across three model families, namely Llama-3.1, Qwen3, and Olmo-3 (both Instruct and Think variants), show consistent improvements, with the magnitude of gains varying across models. 
We find that these differences correlate with the sparsity of the retrieval score distribution: Models with sparser distributions respond strongly to the method, while those with less sparse distributions show relatively modest gains. 
This finding reconfirms the functional importance of retrieval heads in long-context processing from a model development perspective.

The contributions of this work are as follows.
(1) We propose RetMask, a simple and effective method for improving long-context processing that leverages retrieval heads as a source of contrastive training signals, requiring neither human-crafted criteria nor an LLM judge.
(2) We validate the effectiveness of RetMask on four models across three families, where consistent improvements in long-context performance are observed.
We further show that the magnitude of gains correlates with the sparsity of the retrieval score distribution, providing mechanistic insight into when and why the method is effective.
(3) RetMask improves long-context capabilities without degrading general abilities: the trained models maintain performance on mathematics, coding, and general knowledge tasks.

\section{Preliminary: Retrieval Heads}

Prior study has uncovered retrieval heads, a set of attention heads that retrieve relevant information from previous contexts during generation~\cite{wu2025retrieval}.
The algorithm to detect retrieval heads roots from the \textit{Needle-In-A-Haystack} (NIAH) task.

\paragraph{Needle-In-A-Haystack~\cite{niah}.}
For each question $q$ and its corresponding answer $k$ (the ``needle''), the answer $k$ is randomly inserted into a context $x=p_1,\dots,p_n$ composed of $n$ passages that are irrelevant to both $q$ and $k$ (the ``haystack'').
This yields $x'= p_1 \dots k \dots p_n$, where the indices of inserted needle tokens are denoted as $\mathcal{I}_k$.
A language model receives the context with the answer inserted $x'$, along with the question $q$, and is evaluated on whether it correctly outputs $k$.
If successful, the model retrieves the target answer span $k$ from the long context $x'$ by performing a copy-paste operation.

\paragraph{Retrieval Head.}
To detect retrieval heads, \citet{wu2025retrieval} calculates the frequency of an attention head performing copy-paste operations.
Specifically, during decoding, let $y_t$ denote the current token to be generated, and $\bm{a}_t \in \mathbb{R}^{|x'| + t - 1}$ is the attention scores of a head.
The head is considered to be retrieving (i.e., copy-pasting) the token $x_j$ if $y_t = x_j, j=\argmax (\bm{a}_t)$.
If $j \in \mathcal{I}_k$, the head is retrieving a token from the needle.
The retrieval score of head $h$ is thus defined as:
\begin{equation}
    \text{RetrievalScore} (h) = \frac{1}{|\mathcal{T}|}\sum_{(g_h, k) \in \mathcal{T}}\frac{|g_h\cap k|}{|k|},
\end{equation}
where $\mathcal{T}$ is the set of test instances; in each test, $g_h$ denotes the set of all tokens retrieved by the head $h$, and $k$ denotes the needle sequence. 
This metric thus quantifies the overlap between tokens retrieved by the head $h$ and those in the needle sequence.
The scores of all attention heads are computed, and those heads with $\text{RetrievalScore}(h) \ge \tau$ are considered as retrieval heads, where $\tau$ is a threshold hyper-parameter.

\section{Methodology: RetMask}
\begin{figure*}[t]
    \centering
    \includegraphics[width=\linewidth]{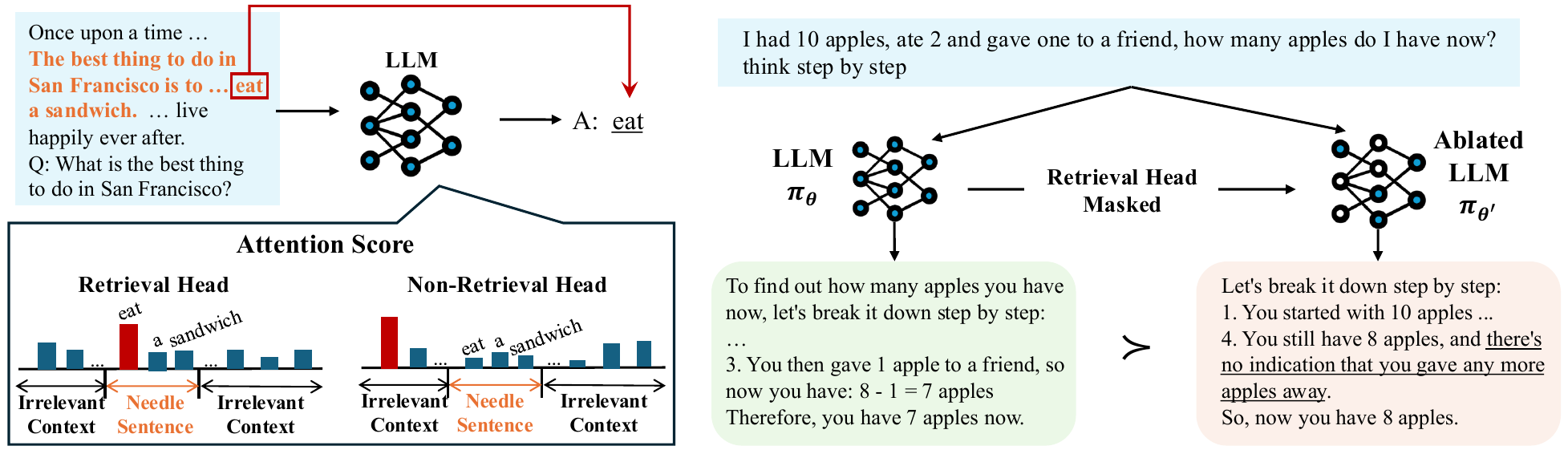}
    \caption{Overview of Preliminaries (left) and RetMask (right). The example on the right is a real case extracted from the training data. We detect and mask retrieval heads for generating contrastive responses. }
    \label{fig:details}
\end{figure*}

This work evaluates the effectiveness of retrieval heads in enhancing the long-context processing capabilities of LLMs (Figure~\ref{fig:details}).
Given an LLM $\pi_{\theta}$, our approach trains the model to prefer outputs sampled from the LLM $\pi_{\theta}$ over those from an ablated variant $\pi_{\theta'}$ (with retrieval heads masked).
The method consists of three stages: (1) Retrieval Head Deactivation; (2) Contrastive Response Generation; (3) Direct Preference Optimization.

\paragraph{Retrieval Head Deactivation.}
Following~\citet{wu2025retrieval}, for a given LLM $\pi_{\theta}$, we compute the retrieval score of all attention heads and detect retrieval heads on top of the NIAH task.
Attention heads with score greater than $\tau$ comprise the retrieval head set $\mathcal{H}_{\text{ret}}$.
We then construct an ablated LLM $\pi_{\theta'}$ by deactivating the identified retrieval heads $\mathcal{H}_{\text{ret}}$.
For each head $h \in \mathcal{H}_{\text{ret}}$, we zero out the corresponding columns in the attention output projection matrix $\bm{W}_o$, thereby preventing the head from contributing to subsequent layers.
\begin{equation}
    \bm{W}_{o'}^h = \begin{cases}
        \bm{0} & \text{if $h \in \mathcal{H}_{\text{ret}}$}\\
        \bm{W}_o^h & \text{otherwise}
    \end{cases}
\end{equation}

\paragraph{Contrastive Response Generation.}
\label{sec:resp_generation}
We synthesize data for direct preference optimization using the model $\pi_{\theta}$ and its ablated variant $\pi_{\theta'}$, shaping contrasts that highlight the contribution of retrieval heads.
To this end, we utilize existing instruction-tuning datasets, which consist of instruction-response pairs.
For each instruction $x$ in the dataset, we discard the original response and generate new responses using $\pi_{\theta}$ and $\pi_{\theta'}$:
\begin{align}
    & y_w \sim \pi_{\theta}(\cdot|x), \label{eq:win} \\
    & y_l \sim \pi_{\theta'}(\cdot|x),  \label{eq:lose}
\end{align}
where Equation~\ref{eq:win} represents sampling from the original LLM, and 
Equation~\ref{eq:lose} represents sampling from the ablated variant.
The response $y_w$ generated under zero perturbation serves as the chosen response, while $y_l$ generated with retrieval heads deactivated serves as the rejected response.
We provide examples about typical failure modes of the rejected responses in Appendix~\ref{appdx:examples}.

\paragraph{Direct Preference Optimization.}

Combining the synthesized responses $y_w, y_l$ with the original instruction $x$, we obtain preference tuples $\{(x, y_w, y_l)\}$.
We train the target policy $\pi_\theta$ using Direct Preference Optimization (DPO,~\citealp{rafailov2023direct}), with the reference policy $\pi_{\text{ref}}$ initialized from the original model.
The objective is:

{\small
\begin{align}
\begin{split}
 & \mathcal{L}(\pi_\theta) =  \\
& {- \mathbb{E} \bigg[\log \sigma \bigg(\beta\log\frac{\pi_\theta(y_w|x)}{\pi_{\text{ref}}(y_w|x)} - \beta\log\frac{\pi_\theta(y_l|x)}{\pi_{\text{ref}}(y_l|x)}\bigg)\bigg],}
\end{split}
\end{align}
}
where $\beta$ is a temperature parameter controlling the deviation from the reference policy.
RetMask uses self-synthesis, i.e., the model used for response generation is the same as the target model, by default; cross-model synthesis results are in Appendix~\ref{appdx:syn_diff}.

\section{Experiments}

\begin{table*}[t]
  \centering
  \small
  \begin{tabular}{lccccc|ccccc}
\Xhline{3\arrayrulewidth}
\textbf{DPO} & \multicolumn{5}{c}{\textbf{Llama-3.1-8B-Instruct}} & \multicolumn{5}{c}{\textbf{Qwen3-8B}}  \\
\textbf{Strategy} & \textbf{8K} & \textbf{16K} & \textbf{32K} & \textbf{64K} & \textbf{128K} & \textbf{8K} & \textbf{16K} & \textbf{32K} & \textbf{64K} & \textbf{128K} \\
\Xhline{2\arrayrulewidth}
-- & 56.03 & 54.14 & 52.42 & 51.65 & 46.40 & 53.20 & 50.16 & 49.89 & 45.44 & 44.73 \\
\hdashline
Smaller-Model & 56.77 & 55.32 & \textbf{53.48} & 52.18 & 47.53 & 52.52 & 49.81 & 48.71 & 46.67 & 45.51 \\
Win-Lose-Pair & 56.50 & 54.42 & 52.47 & 51.62 & 46.05 & 52.80 & 50.14 & 49.71 & 45.93 & 44.49 \\
Non-Retrieval-Mask & 56.45 & 55.55 & 53.19 & 52.14 & 47.19 & 53.02 & 50.28 & 48.67 & \textbf{46.79} & 45.48 \\
Random-Mask & 56.67 & 55.95 & 53.14 & 52.30 & 47.04 & 49.99 & 47.02 & 45.76 & 43.85 & \textbf{45.86} \\
\hline
RetMask & \textbf{58.14} & \textbf{56.92} & \textbf{53.48} & \textbf{53.15} & \textbf{48.68} & \textbf{53.77} & \textbf{50.61} & \textbf{50.34} & \textbf{46.79} & 45.62 \\
\Xhline{3\arrayrulewidth}
\end{tabular}
  \caption{Performance of Llama-3.1 and Qwen3 trained with different strategies on HELMET~\cite{yen2025helmet}.
Models are evaluated using input sequences of 8K, 16K, 32K, 64K, and 128K tokens.
Overall, training with retrieval heads ablated (i.e., RetMask) yields the best performance.}
  \label{tab:main_results}
\end{table*}

\begin{table*}[t]
  \centering
  \small
  \begin{tabular}{lc|ccccccc}
  \Xhline{3\arrayrulewidth}
  \textbf{DPO} & \multicolumn{8}{c}{\textbf{Llama-3.1-8B-Instruct}}  \\
  \textbf{Strategy} & \textbf{Average} & \textbf{Recall} & \textbf{RAG} & \textbf{Cite} & \textbf{Re-rank} & \textbf{ICL} & \textbf{LongQA} & \textbf{Summ} \\
  \Xhline{2\arrayrulewidth}
  -- & 46.40 & 95.13 & 58.58 & 3.09 & 13.73 & 83.80 & 42.69 & 27.81\\
  \hdashline
  Smaller-Model & 47.53 & 94.19 & \textbf{60.83} & 4.22 & 13.44 & 83.76 & 43.15 & 33.12  \\
  Win-Lose-Pair & 46.05 & 93.56 & 59.50 & 3.72 & 12.47 & 83.36 & 39.26 & 30.48\\
  Non-Retrieval-Mask & 47.19 & \textbf{96.69} & 59.00 & 3.45 & 11.38 & 84.28 & 40.93 & \textbf{34.62}\\
  Random-Mask & 47.04 & 96.38 & 59.29 & 3.88 & 10.79 & 83.52 & 41.32 & 34.10 \\
  \hline
  RetMask & \textbf{48.68} & 95.44 & 59.71 & \textbf{5.25} & \textbf{18.16} & \textbf{84.92} & \textbf{43.84} & 33.45\\
  \Xhline{3\arrayrulewidth}
  \end{tabular}
  \caption{Model performance on each task of HELMET when the input sequence length is 128K. The advantage of RetMask is evident on real-world tasks such as generation with citation and passage re-ranking.}
  \label{tab:results_llama}
\end{table*}

\subsection{Settings}

\paragraph{Models.} We evaluate RetMask on three model families: Llama-3.1-8B-Instruct~\cite{grattafiori2024llama3herdmodels}, Qwen3-8B~\cite{yang2025qwen3technicalreport}, and Olmo-3-7B-(Instruct/Think)~\cite{olmo2025olmo3}.
We identify retrieval heads using threshold $\tau=0.1$ for Llama-3.1 and $\tau=0.05$ for Qwen3 and Olmo-3 \footnote{We tuned the hyper-parameter in pilot experiments as explained in Appendix~\ref{appdx:hypara} and \ref{appdx:sweet_spot}.}.
For Qwen3, we disable reasoning when computing the retrieval scores to match the default setting and enable reasoning during contrastive response generation unless otherwise stated.

\paragraph{Benchmarks.} We evaluate on HELMET~\cite{yen2025helmet}, a comprehensive benchmark for long-context processing that covers both synthetic and real-world tasks, categorized as Synthetic Recall (\textit{Recall},~\citealp{hsieh2024ruler}), Retrieved Augmented Generation (\textit{RAG}), Generation with Citations (\textit{Cite}), Passage Re-Ranking (\textit{Re-rank}), Many-Shot In-Context Learning (\textit{ICL}), Long-Document Question Answering (\textit{LongQA}), and Summarization (\textit{Summ}). 
The benchmark covers five context lengths ranging from 8K to 128K tokens.
For Qwen3-8B, we enable reasoning during evaluation.

\paragraph{Retrieval-Head Detection.}  We adopt the script provided by~\citet{wu2025retrieval} that runs NIAH on 20 different context lengths uniformly distributed between 0 and 5K, with the needle inserted at 10 depth positions for each length\footnote{\url{https://github.com/nightdessert/Retrieval_Head}}.
This setting is recommended by the authors, who note that the detection stabilizes with only a few samples. 

\paragraph{Training Data.} RetMask is applicable to any dataset containing user instructions. 
We primarily use LMSYS-Chat-1M~\cite{zheng2024lmsyschatm}, a large-scale collection of human-LLM conversations.
We also experiment with WildChat~\cite{zhao2024wildchat}, another general-purpose dataset collected from human-LLM interactions, in \S~\ref{exp:wildchat}, and Guru~\cite{cheng2025revisiting}, a reinforcement learning dataset, in Appendix~\ref{appdx:rl}.
All training data, with statistics shown in Appendix~\ref{appdx:training_data}, are distinct from the evaluation benchmark,
ensuring that performance gains reflect improvements in long-context capability rather than task-specific tuning.

\paragraph{Baselines.} To focus on the contribution of retrieval heads, we include baselines with different policies of deciding rejected samples $y_l$: 
(1) \textbf{Smaller-Model}: $y_l$ sampled from a smaller LLM, namely Llama-3.2-3B-Instruct~\cite{grattafiori2024llama3herdmodels}\footnote{We experimented with Llama-3.2-1B-Instruct and found the training unstable, thus switched to Llama-3.2-3B-Instruct.} for experiments on Llama-3.1-8B-Instruct and Olmo-3-7B-Instruct, and Qwen3-0.7B for experiments on Qwen3-8B and Olmo-3-7B-Think.
(2) \textbf{Win-Lose-Pair}: $y_l$ sampled from the same LLM but with lower quality. The quality is judged by Gemma-3-27B-IT~\cite{gemmateam2025gemma3technicalreport}.
(3) \textbf{Non-Retrieval-Mask}: $y_l$ sampled from another ablated variant of the LLM, with $|\mathcal{H}_\text{ret}|$ randomly-selected non-retrieval heads masked.
The masked heads are not chosen from the retrieval heads ($h \notin \mathcal{H}_\text{ret}$).
(4) \textbf{Random-Mask}: $y_l$ sampled from another ablated variant of the LLM, with $|\mathcal{H}_\text{ret}|$ heads randomly masked.
Masked heads can be the retrieval heads.
As a strong baseline, we also compare with existing work~\cite{zhang-etal-2025-longreward} in Section~\ref{exp:strong_baselines}.

\subsection{Main Results}
\label{exp:main}

The performance of Llama-3.1-8B-Instruct and Qwen3-8B trained under different strategies is shown in Table~\ref{tab:main_results}.
Additionally, Table~\ref{tab:results_llama} presents per-task performance of Llama-3.1 on HELMET evaluated using input sequences of 128K tokens.
The task-wise performance of Qwen3-8B is detailed in Appendix~\ref{appdx:qwen}.
In all tables throughout this paper, the row labeled as `--' denotes the baseline model before training.

\paragraph{Strong Improvements on Llama-3.1 across all context lengths.}
Table~\ref{tab:main_results} shows that RetMask achieves the best performance across all context lengths when training Llama-3.1.
At 128K, the proposed method improves the base model by 2.28 points ($46.40 \to 48.68$).
The improvement persists across context lengths ranging from 8K to 128K tokens, demonstrating the robustness of the method.
RetMask outperforms the other baselines (Non-Retrieval-Mask and Random-Mask), confirming that improvements stem specifically from 
targeting retrieval heads rather than the ablating operation itself.
Notably, Win-Lose-Pair, which trains the model to prefer 
higher-quality outputs over lower-quality ones from the same model, 
shows decreased performance ($46.40 \to 46.05$).
This indicates that the gains from the proposed method are not 
simply due to preference optimization on output quality, but rather 
from the contrast that specifically targets retrieval capabilities.
We also verified that supervised fine-tuning with chosen responses yields suboptimal performance, as detailed in Appendix~\ref{appdx:sft}.

Interestingly, the training sequences average only 63.62 tokens for inputs and 
494.69 tokens for outputs, significantly shorter than the evaluation 
contexts.
This reveals an advantage of the proposed method: it 
enhances long-context capabilities through short-sequence training, 
consistent with findings in~\citet{gao-etal-2025-train} that post-training with short-context instruction datasets is sufficient for achieving 
good long-context performance.

\paragraph{Improvements on Qwen3 across all context lengths.}
On Qwen3, consistent improvements are also observed with RetMask: +0.57 at 8K, +0.45 at 16K, +0.45 at 32K, +1.35 at 64K, +0.89 at 128K.
However, the improvements are modest compared to those on Llama-3.1, and the Random-Mask baseline was slightly better than RetMask by 0.24 points when the input sequence is 128K tokens long.
We attribute this to the fundamental differences in retrieval score distribution as detailed in \S~\ref{sec:distribution}.
We also provide a detailed analysis of how Qwen3's reasoning contents affect the performance in \S~\ref{exp:reasoning}.

\paragraph{Improvement is significant on tasks requiring long-context processing.}
Table~\ref{tab:results_llama} details the task-specific impact of the method on Llama-3.1 evaluated using input sequences of 128K tokens.
We observe particularly significant improvements on tasks requiring precise information retrieval:
\textit{Cite} improved from 3.09 to 5.25 (70\% relative improvement) and \textit{Re-rank} improved from 13.73 to 18.16 (32\% relative improvement).
Both tasks require referring back to the document segments in context and generating text while reorganizing them.
These results validate that strengthening retrieval heads enhances both the model's ability to locate information in long contexts and its capacity to generate well-grounded, context-backed responses.

\subsection{Comparison with Existing Methods}
\label{exp:strong_baselines}

\begin{table}[t]
  \centering
  \small
    \begin{tabular}{lccccc}
    \Xhline{3\arrayrulewidth}
    \textbf{} & \textbf{8K} & \textbf{16K} & \textbf{32K} & \textbf{64K} & \textbf{128K}\\
    \Xhline{2\arrayrulewidth}
    -- & 56.03 & 54.14 & 52.42 & 51.65 & 46.40 \\
    \hdashline
    LongReward & 56.53 & 54.74 & 52.50 & 52.14 & 46.71 \\
    RetMask$^{*}$ & 57.21 & 55.31 & 52.87 & 51.97 & 46.89 \\
    RetMask & \textbf{58.14} & \textbf{56.92} & \textbf{53.48} & \textbf{53.15} & \textbf{48.68} \\ 
\Xhline{3\arrayrulewidth}
    \end{tabular}
\caption{Comparison of Llama-3.1-8B-Instruct fine-tuned via DPO using 
LongReward (\citet{zhang-etal-2025-longreward}, the existing method) and 
RetMask (the proposed method), evaluated on HELMET across input lengths 
of 8K--128K tokens. RetMask$^{*}$ is a downsampled variant of RetMask 
matched to LongReward's sample size. Even with downsampling, RetMask 
consistently outperforms LongReward.}
  \label{tab:strong_baseline}
\end{table}

\begin{table*}[t]
  \centering
  \small
    \begin{tabular}{lcccc|cccc}
    \Xhline{3\arrayrulewidth}
    \textbf{DPO} & \multicolumn{4}{c}{\textbf{Olmo-3-7B-Instruct}} &
    \multicolumn{4}{c}{\textbf{Olmo-3-7B-Think}}\\
    \textbf{Strategy} & \textbf{8K} & \textbf{16K} & \textbf{32K} & \textbf{64K} &
    \textbf{8K} & \textbf{16K} & \textbf{32K} & \textbf{64K}\\
    \Xhline{2\arrayrulewidth}
    -- & 43.73 & 40.09 & 33.21 & 25.00 & 46.53 & 45.83 & 42.41 & 35.07  \\
    \hdashline
    Smaller-Model & 41.96 & 37.11 & 30.63 & 22.02 & 45.26 & 44.44 & 42.60 & 33.92  \\
    Non-Retrieval-Mask & 42.91 & 39.24 & 32.24 & 23.18 & 46.46 & 45.56 & 43.04 & 34.34  \\ 
    \hline
    RetMask  & \textbf{45.51} & \textbf{41.75} & \textbf{34.28} & \textbf{25.59} & \textbf{46.69} & \textbf{46.07} & \textbf{43.09} & \textbf{35.54}
\\
\Xhline{3\arrayrulewidth}
    \end{tabular}
  \caption{Performance of Olmo-3 models fine-tuned via DPO with different 
strategies, evaluated on HELMET across input lengths of 8K--64K tokens. 
RetMask yields pronounced gains on both the Instruct and the 
Think variant.}
  \label{tab:olmo}
\end{table*}

To situate RetMask within the broader landscape of approaches for improving LLMs' long-context capabilities, we compare it against LongReward~\cite{zhang-etal-2025-longreward}, a recent DPO method that leverages AI feedback for long-context processing.
Specifically, we fine-tune Llama-3.1-8B-Instruct via DPO on data generated by each method and evaluate the resulting models on HELMET. 
For LongReward, we use the officially released dataset\footnote{\textit{dpo\_llama3.1\_8b} split of \url{https://huggingface.co/datasets/zai-org/LongReward-10k}}.
Results are reported in Table~\ref{tab:strong_baseline}.

\paragraph{RetMask more effectively improves the long-context processing ability than LongReward.}
While training on LongReward yields gains over the untrained baseline, these improvements clearly lag behind those achieved by training on RetMask.
One might attribute the gap to RetMask's larger dataset size 
(LongReward: 10K samples vs.\ RetMask: 294K samples).
To isolate the effect of the method from that of dataset size, we train Llama-3.1-8B-Instruct on a varient of the RetMask-based dataset that is downsampled to match the sample size of the LongReward-based dataset.
The downsampled variant performs below the full RetMask, but still 
consistently outperforms the LongReward-trained model.
We therefore conclude that RetMask is a more effective approach to improving 
long-context processing, independent of dataset size.
Unlike LongReward, which relies on an LLM judge to score responses based on 
human-crafted criteria, RetMask automatically treats responses from 
retrieval-head-masked models as rejected, and those from the original model 
as chosen.
This eliminates the need for both human-crafted criteria and LLM judges, yet training on RetMask still outperforms training on LongReward, which underscores the power of grounding contrastive signals in mechanistic interpretability.

While the dominant approach to extending context length involves continual 
pre-training~\cite{grattafiori2024llama3herdmodels,yang2025qwen3technicalreport,olmo2025olmo3,wu-etal-2025-longattn}, we note that RetMask is complementary to this line of work: It is possible to first extend context length via continual pre-training, then apply RetMask as an additional post-training stage to further improve long-context performance without compromising general capabilities (\S~\ref{ana:others}).
We thus conclude that RetMask is not only of academic interest but also of practical value as a lightweight addition to the standard model development pipeline for long-context LLMs.

\subsection{Generalization Across Alignment Objectives}
\label{exp:olmo}

Having witnessed the effectiveness of RetMask on Llama-3.1 and Qwen3, we now experiment on the Olmo-3 family, which comprises a standard instruction-tuned variant (Olmo-3-7B-Instruct) and a reasoning-focused variant (Olmo-3-7B-Think). 
This allows us to examine how RetMask generalizes across models with different alignment objectives while controlling for pre-training and mid-training processes.
We include two of the strongest baselines: DPO with a weaker model and arbitrary non-retrieval attention heads masked.
As Olmo-3's maximum content length is 64K, we evaluate on input lengths up to 64K and exclude the 128K setting.
Results are shown in Table~\ref{tab:olmo}. 

\paragraph{RetMask improves over both variants.}
Consistent with results on Llama-3.1 and Qwen3, RetMask yields clear performance gains over all baselines on both Olmo-3-7B-Instruct and Olmo-3-7B-Think across all input lengths.
This confirms that the effectiveness of RetMask generalizes across different alignment objectives.
Notably, the gains are more pronounced on the Instruct variant than on the 
Think variant.
The reasons may be manifold, with one possible explanation concerning the retrieval head detection process: while NIAH assumes that a model directly outputs the answer, Olmo-3-7B-Think first generates the reasoning contents before the answer, which may degrade the accuracy of retrieval head detection.
We leave a deeper investigation into the underlying reasons for future work.

\subsection{Robustness with Reasoning Mode}
\label{exp:reasoning}
\begin{figure}
    \centering
    \includegraphics[width=\linewidth]{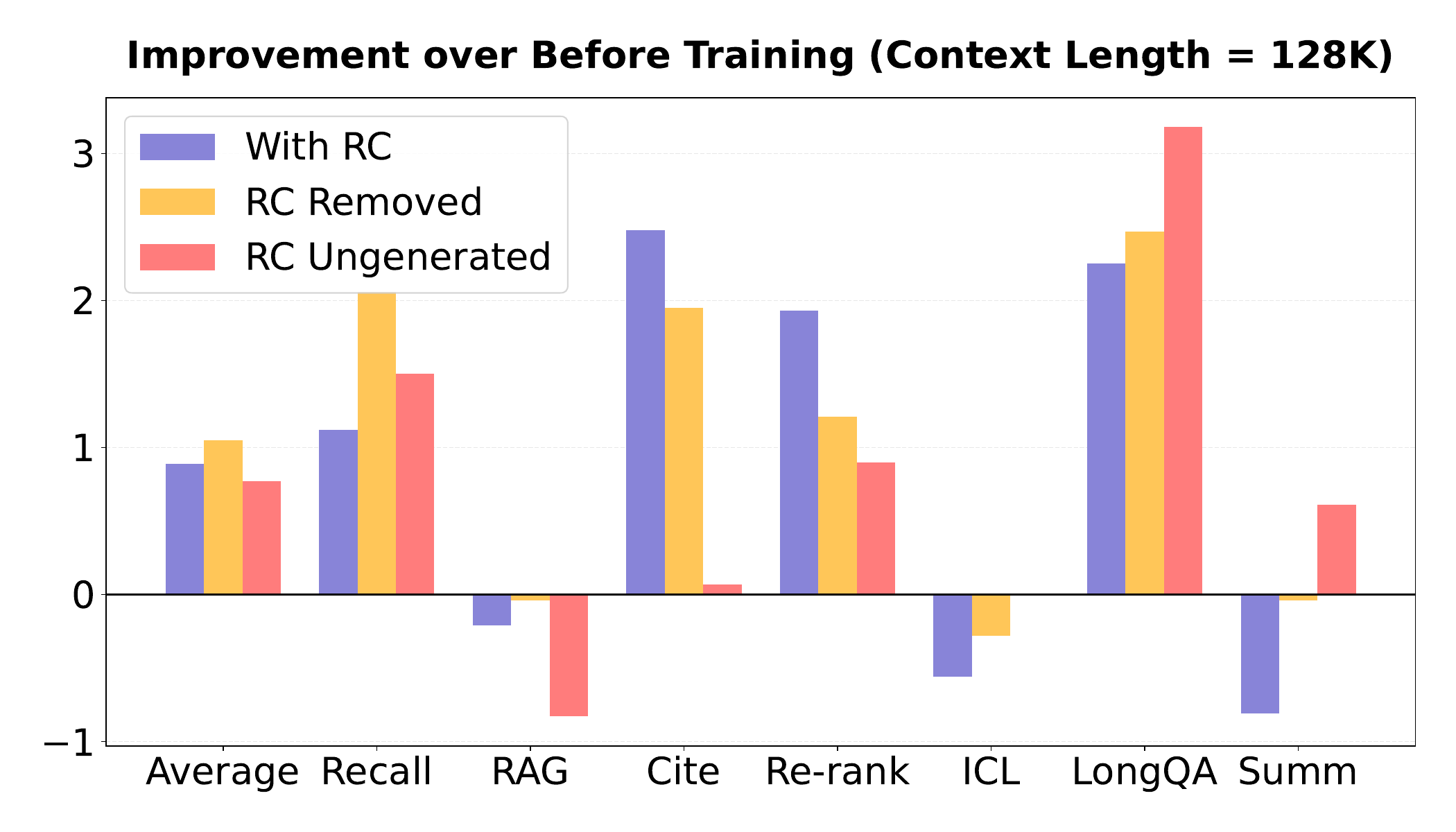}
    \caption{Per-task performance gains of Qwen3-8B on HELMET at 128K input tokens. RC denotes reasoning contents. RetMask consistently improves over the checkpoint before training, regardless of whether reasoning contents are included in the training samples.}
    \label{fig:rc}
\end{figure}

\begin{table*}[t]
  \centering
  \small
  \begin{tabular}{lc|ccccccc}
  \Xhline{3\arrayrulewidth}
  \textbf{DPO} & \multicolumn{8}{c}{\textbf{Llama-3.1-8B-Instruct}}  \\
  \textbf{Strategy} & \textbf{Average} & \textbf{Recall} & \textbf{RAG} & \textbf{Cite} & \textbf{Re-rank} & \textbf{ICL} & \textbf{LongQA} & \textbf{Summ} \\
  \Xhline{2\arrayrulewidth}
  -- & 46.40 & 95.13 & 58.58 & 3.09 & 13.73 & 83.80 & 42.69 & 27.81\\
  \hdashline
  Smaller-Model & 47.30 & 93.56 & \textbf{60.79} & 3.62 & 15.29 & 83.44 & \textbf{42.78} & 31.63 \\
  Win-Lose-Pair & 46.91 & 94.44 & 59.04 & 4.30 & 14.17 & 83.96 & 40.64 & 31.80\\
  Non-Retrieval-Mask & 47.34 & 96.75 & 59.58 & 4.13 & 12.86 & 83.68 & 39.69 & \textbf{34.76} \\
  Random-Mask & 47.23 & \textbf{96.38} & 60.04 & 3.45 & 12.75 & 83.24 & 41.59 & 33.16 \\
  \hline
  RetMask & \textbf{48.83} & 95.81 & 59.63 & \textbf{6.10} & \textbf{19.27} & \textbf{85.32} & 41.87 & 33.83\\
  \Xhline{3\arrayrulewidth}
  \end{tabular}
  \caption{Model performance of Llama-3.1 trained with different strategies using WildChat, evaluated on HELMET when the input sequence length is 128K. The model trained with RetMask scores the highest among all strategies.}
  \label{tab:wildchat}
\end{table*}

Experiments on Qwen3 in \S~\ref{exp:main} use responses generated with reasoning enabled, in which the model outputs reasoning content before producing the final answers.
In this section, we investigate how the reasoning process affects training effectiveness.
To this end, we conduct additional experiments: (1) \textbf{RC removed}: Generate training data with reasoning enabled, then remove the reasoning contents and keep the response only; (2) \textbf{RC Ungenerated}: Generate training data with reasoning disabled.
The trained models are evaluated with reasoning enabled to ensure results are comparable with those in Table~\ref{tab:main_results}.

\paragraph{Removing reasoning contents has minimal impact on the effectiveness of the proposed method.}
Figure~\ref{fig:rc} shows that removing reasoning contents has minimal impact:
Five of seven tasks achieve comparable or better performance than training with full reasoning contents.
This indicates that ablating retrieval heads degrades response quality sufficiently to provide effective DPO training signals, even without reasoning contents in the training samples.
Thus, our method's core mechanism, retrieval head ablation, drives improvements regardless of whether reasoning contents are preserved or not.

\paragraph{Reasoning contents are important for complex tasks.}
Performance of tasks requiring complex reasoning, namely \textit{Cite} and \textit{Re-rank},
degrades significantly when trained with reasoning contents removed or ungenerated (Figure~\ref{fig:rc}).
This demonstrates that for tasks involving source tracking and passage comparison, explicit reasoning chains in training data are important for RetMask to achieve optimal effectiveness:
The reasoning content helps the model learn not just retrieval patterns, but also how to reason over retrieved information.
For such complex tasks, preserving reasoning contents during training ensures the effectiveness of RetMask.

\subsection{Robustness Across Training Datasets}
\label{exp:wildchat}

\S~\ref{exp:main} has demonstrated the effectiveness of RetMask in enhancing long-context capabilities using LMSYS-Chat-1M~\cite{zheng2024lmsyschatm}.
A potential concern is whether the improvements stem from the retrieval-ablated optimization strategy or from dataset-specific characteristics that happen to enhance long-context processing.
To address this, we conduct experiments using Wildchat~\cite{zhao2024wildchat}, another dataset collected for instruction tuning.
Specifically, we synthesize responses to first-turn instructions in WildChat using RetMask to build the training data.
Evaluation results of the trained models are shown in Table~\ref{tab:wildchat}.
\paragraph{Improvements are consistent on WildChat.}
As with LMSYS-Chat-1M, RetMask outperforms all baselines on average across tasks.
Notably, substantial improvements are observed on \textit{Cite} and \textit{Re-rank} tasks, consistent with findings in \S~\ref{exp:main}.
These results confirm that the improvements are attributed to the optimization methodology rather than dataset-specific artifacts.
This demonstrates the robustness and generalizability of RetMask across different datasets.

\section{Analysis}
\label{sec:analysis}

\subsection{Performance on Other Tasks}
\label{ana:others}

\begin{table}[t]
  \centering
  \small
  \begin{tabular}{lccccc}
  \Xhline{3\arrayrulewidth} 
  & \textbf{MTB} & \textbf{GPQA} & \textbf{MATH} & \textbf{HE} & \textbf{MMLUP}\\
  \Xhline{2\arrayrulewidth}
  \multicolumn{6}{l}{\textbf{(a) Llama-3.1-8B-Instruct}} \\
  Before & 0.75 &	0.25 &	\textbf{0.53} &  \textbf{0.71} & \textbf{0.49} \\
  After & \textbf{0.77} &	\textbf{0.33} & 0.52 &	0.68 & 0.48 \\
  \hline
  \multicolumn{6}{l}{\textbf{(b) Qwen3-8B}} \\
  Before & 0.86 & 0.56 & \textbf{0.97} & 0.89 & 0.71\\
  After &  \textbf{0.88} & \textbf{0.60} & \textbf{0.97} & \textbf{0.92} & \textbf{0.74}\\
  \hline
  \multicolumn{6}{l}{\textbf{(c) Olmo-3-7B-Instruct}} \\
  Before & \textbf{0.81} & 0.36 & \textbf{0.87} & \textbf{0.82} & \textbf{0.59}\\
  After & \textbf{0.81} & \textbf{0.44} & 0.83 & 0.81 & 0.58 \\  
  \hline
  \multicolumn{6}{l}{\textbf{(d) Olmo-3-7B-Think}} \\
  Before & 0.62  & \textbf{0.52} & 0.95 & 0.92 & 0.62 \\
  After & \textbf{0.63} & \textbf{0.52} & \textbf{0.96} & \textbf{0.94} & \textbf{0.63} \\  
  \Xhline{3\arrayrulewidth}
  \end{tabular}
  \caption{Model performance before and after training with RetMask. \textbf{MTB},\textbf{MATH},\textbf{HE}, \textbf{MMLUP} stands for MT-Bench, MATH-500, HumanEval, and MMLU-Pro, respectively. In general, training with RetMask does not degrade the performance on these tasks.}
  \label{tab:other_tasks}
\end{table}

\begin{figure*}[t]
    \centering
    \includegraphics[width=\linewidth]{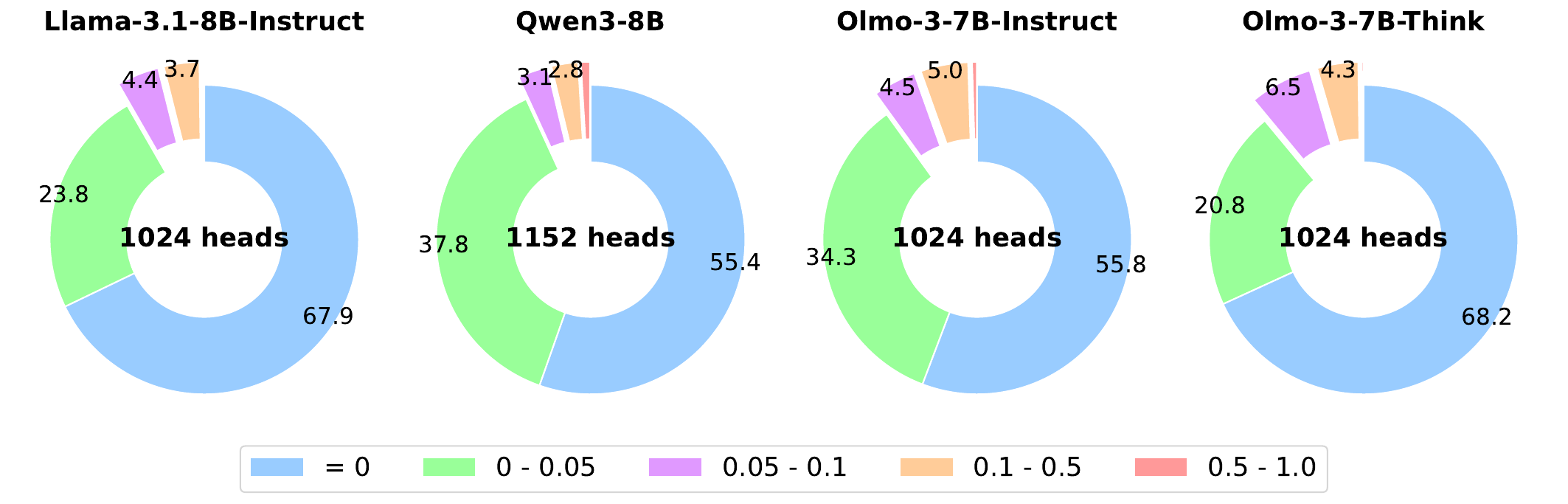}
    \caption{The retrieval score distribution of LLMs tested in this work. While attention heads in Llama-3.1-8B-Instruct exhibit a concentrated pattern of retrieval capabilities, it is more distributed for Olmo-3-7B-Instruct.}
    \label{fig:distribution}
\end{figure*}

The previous section showed that RetMask improves long-context processing. 
A natural question is whether these gains come at the expense of general language understanding and reasoning.
To address this concern, we evaluate trained models on five established benchmarks widely used to assess model capability during LLM development: (1) \textbf{MT-Bench}~\citep{zheng2023judging}: Multi-turn conversational ability;
(2) \textbf{GPQA-Diamond}~\citep{rein2024gpqa}: Expert-level scientific reasoning;
(3) \textbf{MATH-500}~\citep{lightman2023lets}: Mathematical problem-solving; (4) \textbf{HumanEval}~\citep{chen2021evaluating}: Code generation; (5) \textbf{MMLU-Pro}~\citep{wang2024mmlupro}: Broad knowledge and understanding.
%These benchmarks are widely used to assess model capability during LLM development.
Results are presented in Table~\ref{tab:other_tasks}.

\paragraph{RetMask preserves general capabilities.}
Overall, training with RetMask largely preserves general capabilities, with most scores remaining at or above the level before training.
In several cases, we observe modest gains, most notably on GPQA-Diamond.
However, we do not attribute these specifically to the retrieval-head-ablated contrastive signals, as they may reflect general effects of DPO training.
The primary takeaway is that RetMask's long-context improvements do not come at the cost of general language understanding or reasoning.
% We thus conclude that the retrieval-ablated optimization strategy improves the long-context capability without compromising on general abilities.

\subsection{Retrieval Score Distribution}
\label{sec:distribution}

\begin{figure*}
    \centering
    \begin{subfigure}[t]{0.47\textwidth}
    \centering
    \includegraphics[width=1.0\textwidth]{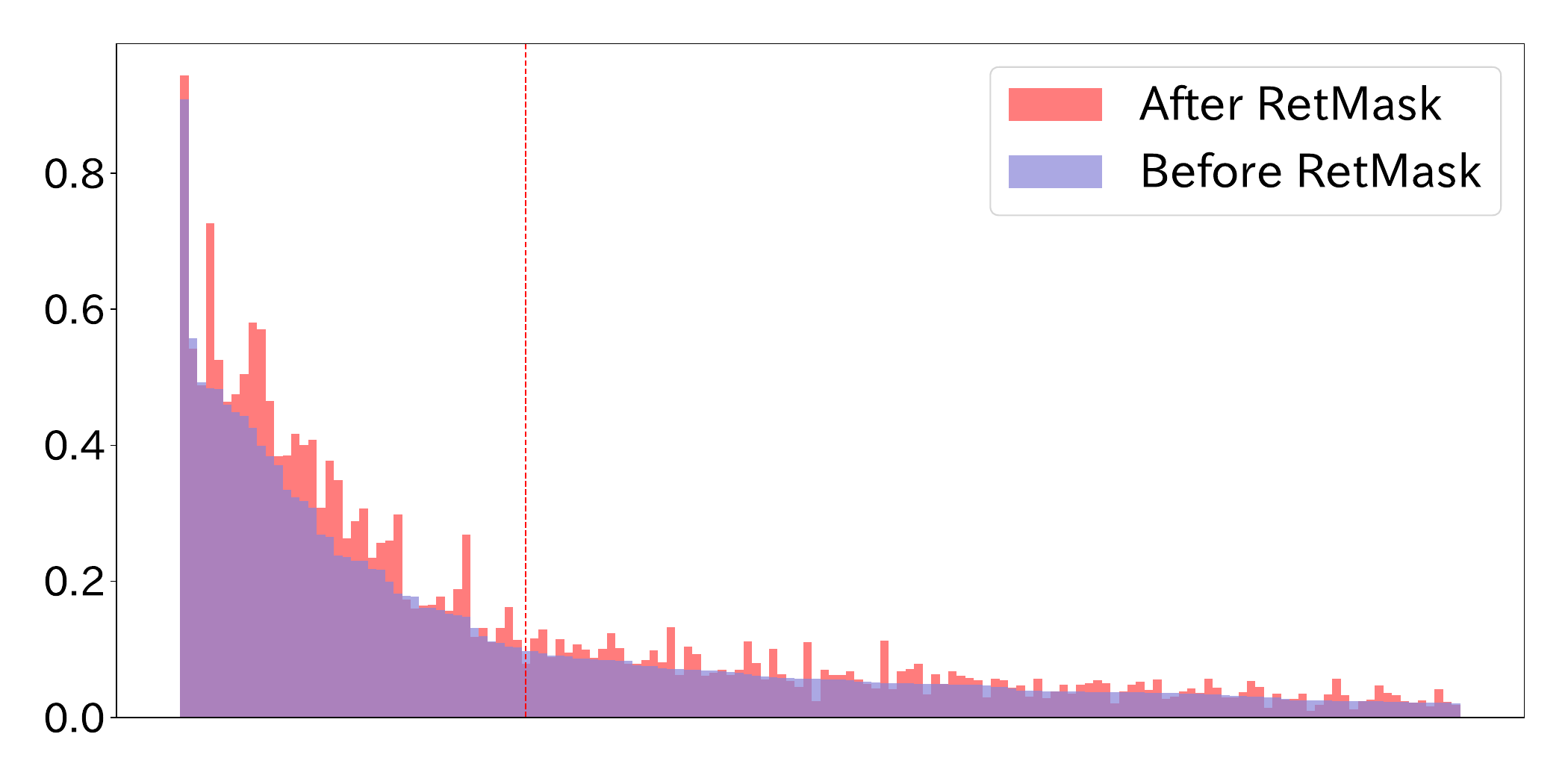}
    \caption{Llama-3.1-8B-Instruct.}
    \end{subfigure}
    \begin{subfigure}[t]{0.47\textwidth}
    \centering
    \includegraphics[width=1.0\textwidth]{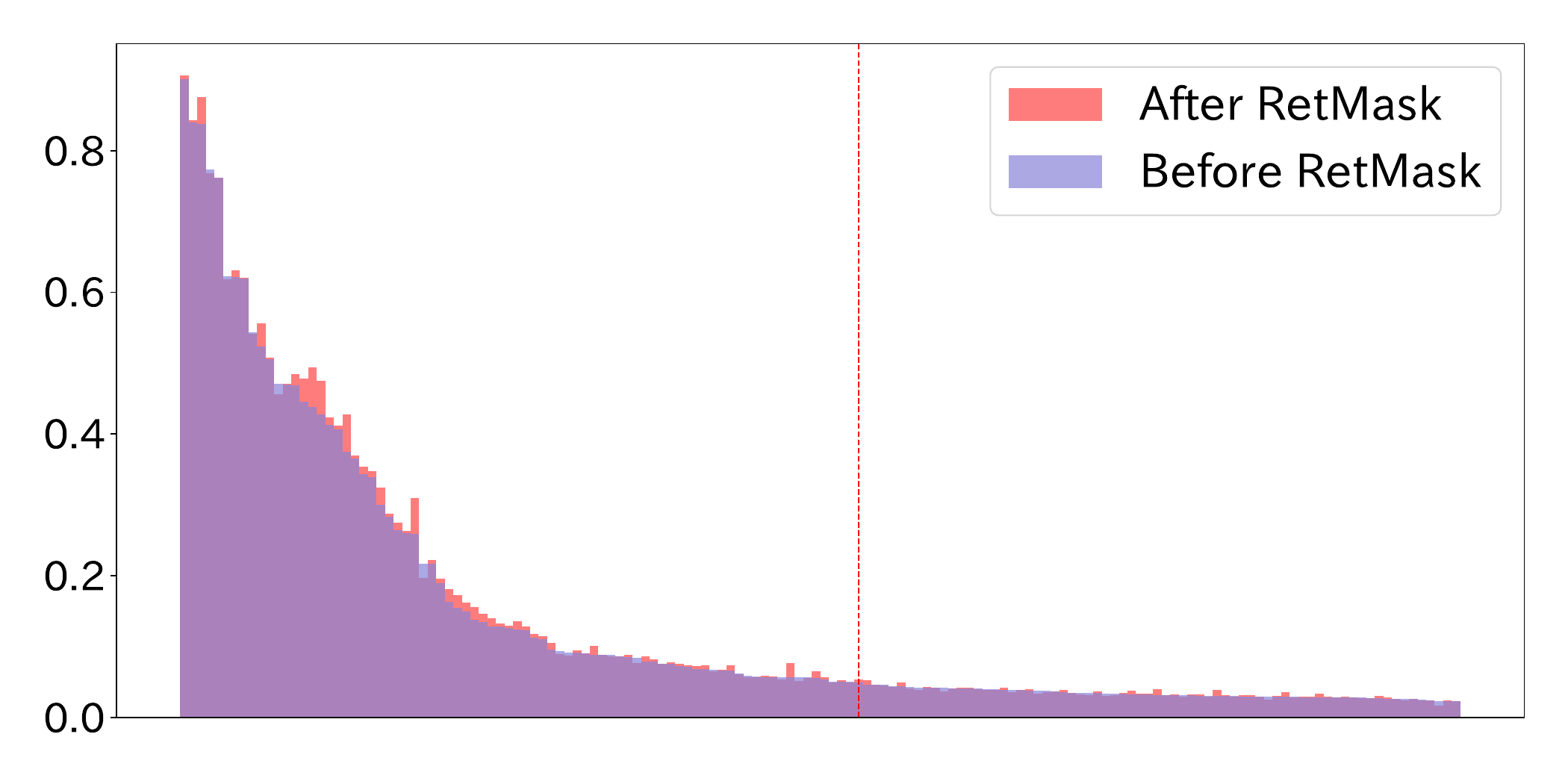}
    \caption{Qwen3-8B.}
    \end{subfigure}
    \caption{The distribution of retrieval score before and after RetMask. We observe an increase in retrieval scores for both Llama-3.1-8B-Instruct and Qwen3-8B. }
    \label{fig:retrieval_score}
\end{figure*}

\S~\ref{exp:main} and \S~\ref{exp:olmo} demonstrate that RetMask achieves the largest improvements on Llama-3.1-8B-Instruct, followed by Olmo-3-7B-Instruct, Qwen3-8B, and Olmo-3-7B-Think.
Here, we explore how the organization of retrieval capabilities across attention heads relates to the effectiveness of RetMask.
To this end, we plot the retrieval score distributions of all four models in Figure~\ref{fig:distribution}\footnote{The distribution of Olmo-3-7B-Think is not directly comparable to the other three, as it generates reasoning contents before the final answer during retrieval score computation.}.

\paragraph{Retrieval capabilities concentrate on a small set of attention heads.}
As in Figure~\ref{fig:distribution}, across all models, only a small proportion of attention heads show high retrieval scores.
This observation is consistent with \citet{wu2025retrieval}, who report a similar distribution of retrieval scores across training stages (pre-trained vs.\ post-trained), architectures (dense vs.\ mixture-of-experts), and parameter sizes.
With thresholds of $\tau \ge 0.1$ for Llama-3.1 and $\tau \ge 0.05$ for Qwen3 and Olmo-3, RetMask masks only 4--10\% of attention heads.

\paragraph{RetMask's effectiveness correlates with the sparsity of the retrieval score distribution.}
This distribution pattern directly influences the effectiveness of RetMask: 
When the retrieval score distribution is sparse, i.e., with retrieval capabilities concentrated in a small subset of heads as in Llama-3.1-8B-Instruct, masking the top-scored heads creates a large performance gap between the original and ablated models, yielding strong contrastive training signals.
Conversely, when the distribution is less sparse, as in Olmo-3-7B-Instruct and Qwen3-8B, the remaining unmasked heads collectively compensate for the ablated ones, reducing the contrast between chosen and rejected responses and thereby 
weakening the training signal.
% These findings validate our hypothesis:
% RetMask's effectiveness depends on the concentration of retrieval capability,
% with models exhibiting specialized retrieval heads responding more strongly to our optimization approach than models with distributed retrieval patterns.
This suggests that the sparsity of the retrieval score distribution may serve as a practical predictor of RetMask's effectiveness, allowing practitioners to gauge applicability before committing to training.

\subsection{RetMask's Effect on Retrieval Heads}
\label{ana:retrieval}

In this section, we analyze how RetMask affects the model by examining changes in the retrieval scores.
Figure~\ref{fig:retrieval_score} displays the retrieval scores of the top 150 heads before training, and how their scores change after training.
The red vertical dashed lines present the masking threshold: heads to the left were masked when generating rejected responses (40 heads for Llama-3.1 and 79 heads for Qwen3).

\paragraph{Retrieval scores improve after training.}
For Llama-3.1-8B-Instruct, we observe clear improvements in retrieval scores after RetMask training.
The average retrieval score increases from 0.017 to 0.020, showing a 17.6\% relative improvement.
For Qwen3-8B, the average score increases from 0.020 to 0.021 (+5\%), reflecting its limited responsiveness to retrieval optimization.
%These enhancements validate the properness of RetMask: Training with the contrastive pairs strengthens the attention heads responsible for long-context retrieval.

\paragraph{Enhancements concentrate on masked heads.}
The improvements are not uniform across all heads.
Specifically, the masked heads show substantial gains, while the other heads exhibit minor changes.
For Llama-3.1, the masked heads exhibit an average improvement of 0.051, while non-masked heads show modest changes (average +0.001).
This demonstrates that RetMask selectively strengthens the retrieval heads targeted during training.

\section{Related Work}

\paragraph{Long-Context Language Modeling.} 
Existing methods for long-context LLMs focus on data engineering.
Common approaches include adjusting RoPE frequency and staged continual pre-training~\cite{grattafiori2024llama3herdmodels,yang2025qwen3technicalreport,gao-etal-2025-train}.
For instance, \citet{grattafiori2024llama3herdmodels} extends context windows over five stages, and \citet{gao-etal-2025-train} seeks an optimal mix of short and long context data during multi-stage training.
For post-training, findings are mixed: \citet{bai-etal-2024-longalign} reports benefits from long-context fine-tuning, while~\citet{gao-etal-2025-train} finds short sequences sufficient.
Closer to our work, \citet{wu-etal-2025-longattn} also leverages attention patterns, but focuses on data selection based on dependency distances.
All these studies emphasize data, whereas we take a model-centric approach through mechanistic interpretability.

\paragraph{Mechanistic Interpretability of LLMs.}
Studies have been conducted to understand the functionality of LLM components. 
\citet{meng2022locating} has revealed that knowledge can be located and edited by manipulating specific neurons.
\citet{tang-etal-2024-language} and \citet{hiraoka-inui-2025-repetition} have demonstrated the existence of language-specific and repetition neurons, respectively.
%These studies are conducted during inference time, when researchers activate or deactivate the neurons and study models' behaviors.
However, few studies connect the discovery to the development of better models. \citet{mondal-etal-2025-language} reported that language-specific neurons cannot facilitate cross-lingual transfer.
Our work takes a step toward connecting mechanistic interpretability with the development of LLMs, specifically by leveraging it to build more effective models within the context of long-context processing.
Closest to ours, \citet{huang-etal-2025-improving} applies retrieval-head masking to improve the faithfulness of long-form question answering via SFT; we instead target general long-context capabilities with DPO and link effectiveness to retrieval-score sparsity.
% This, in turn, provides evidence for the existence of neural components, i.e., the retrieval head in this study.

\section{Conclusion}

This work explores how mechanistic interpretability can facilitate model development in the context of long-term context processing.
By collecting contrastive response pairs through selective deactivation of retrieval heads, we develop RetMask, an approach that enhances long-context capabilities without compromising general capabilities.
Experiments on four models across three families demonstrate consistent improvements, with gains correlating with the sparsity of the retrieval score distribution: Models with sparser distributions achieve stronger gains, while those with less sparse distributions show more modest improvements.
This systematic relationship between retrieval head sparsity and training effectiveness reconfirms the functional importance of retrieval heads and demonstrates that mechanistic insights can be transformed into tangible performance improvements.

Future work includes investigating scaling to larger models,
developing a theoretical understanding of the underlying mechanisms, and extending this approach to other specialized components.

\section*{Limitations}

This work focuses on models up to 8B parameters.
Scaling to larger models remains an open question, though \citet{wu2025retrieval} 
report that retrieval head organization patterns persist at larger scales, 
suggesting that the core mechanism should generalize.
A second limitation concerns retrieval head detection: we rely on NIAH, a synthetic task that targets copy-paste retrieval mechanisms.
Exploring detection methods grounded in real-world data is a promising direction that could further improve effectiveness, and we leave this to future work.

\section*{Ethics Considerations}

\paragraph{Data and Safety.}
We use publicly available datasets (LMSYS-Chat-1M~\cite{zheng2024lmsyschatm}, WildChat~\cite{zhao2024wildchat}, Guru-RL-92K~\cite{cheng2025revisiting}) 
with standard filtering for toxic content and personally identifiable information.
However, some potentially harmful content may remain.
Models trained with our method should undergo standard safety alignment before deployment.

\paragraph{Synthetic Generation.}
Our method generates synthetic training data by contrasting full and retrieval-ablated model outputs, without introducing new information about real individuals.

\section*{Acknowledgments}

We thank the reviewers and area chairs for their feedback. 
We corrected an implementation error in the Olmo experiments, which revealed that RetMask is effective across all tested models.

This work was supported by JST ACT-X, Japan, Grant Number JPMJAX25CN.
This work was also supported by JSPS KAKENHI Grant Number 25H01137.
Experiments were carried out using the TSUBAME4.0 supercomputer at Institute of Science Tokyo.
We also used ABCI 3.0 provided by AIST and AIST Solutions with support from ``ABCI 3.0 Development Acceleration Use''.

% This document has been adapted
% by Steven Bethard, Ryan Cotterell and Rui Yan
% from the instructions for earlier ACL and NAACL proceedings, including those for
% ACL 2019 by Douwe Kiela and Ivan Vuli\'{c},
% NAACL 2019 by Stephanie Lukin and Alla Roskovskaya,
% ACL 2018 by Shay Cohen, Kevin Gimpel, and Wei Lu,
% NAACL 2018 by Margaret Mitchell and Stephanie Lukin,
% Bib\TeX{} suggestions for (NA)ACL 2017/2018 from Jason Eisner,
% ACL 2017 by Dan Gildea and Min-Yen Kan,
% NAACL 2017 by Margaret Mitchell,
% ACL 2012 by Maggie Li and Michael White,
% ACL 2010 by Jing-Shin Chang and Philipp Koehn,
% ACL 2008 by Johanna D. Moore, Simone Teufel, James Allan, and Sadaoki Furui,
% ACL 2005 by Hwee Tou Ng and Kemal Oflazer,
% ACL 2002 by Eugene Charniak and Dekang Lin,
% and earlier ACL and EACL formats written by several people, including
% John Chen, Henry S. Thompson and Donald Walker.
% Additional elements were taken from the formatting instructions of the \emph{International Joint Conference on Artificial Intelligence} and the \emph{Conference on Computer Vision and Pattern Recognition}.

\bibliography{custom}

@inproceedings{
wu2025retrieval,
title={Retrieval Head Mechanistically Explains Long-Context Factuality},
author={Wenhao Wu and Yizhong Wang and Guangxuan Xiao and Hao Peng and Yao Fu},
booktitle={The Thirteenth International Conference on Learning Representations (ICLR)},
year={2025},
url={https://openreview.net/forum?id=EytBpUGB1Z}
}

@inproceedings{
yen2025helmet,
title={{HELMET}: How to Evaluate Long-context Models Effectively and Thoroughly},
author={Howard Yen and Tianyu Gao and Minmin Hou and Ke Ding and Daniel Fleischer and Peter Izsak and Moshe Wasserblat and Danqi Chen},
booktitle={The Thirteenth International Conference on Learning Representations (ICLR)},
year={2025},
url={https://openreview.net/forum?id=293V3bJbmE}
}

@misc{niah,
    title={Needle in a haystack - pressure testing {LLMs}},
    author={Greg Kamradt},
    year={2023},
    url={https://github.com/gkamradt/LLMTest_NeedleInAHaystack},
}

@misc{brown2020languagemodelsfewshotlearners,
      title={Language Models are Few-Shot Learners}, 
      author={Tom B. Brown and Benjamin Mann and Nick Ryder and Melanie Subbiah and Jared Kaplan and Prafulla Dhariwal and Arvind Neelakantan and Pranav Shyam and Girish Sastry and Amanda Askell and Sandhini Agarwal and Ariel Herbert-Voss and Gretchen Krueger and Tom Henighan and Rewon Child and Aditya Ramesh and Daniel M. Ziegler and Jeffrey Wu and Clemens Winter and Christopher Hesse and Mark Chen and Eric Sigler and Mateusz Litwin and Scott Gray and Benjamin Chess and Jack Clark and Christopher Berner and Sam McCandlish and Alec Radford and Ilya Sutskever and Dario Amodei},
      year={2020},
      eprint={2005.14165},
      archivePrefix={arXiv},
      primaryClass={cs.CL},
      url={https://arxiv.org/abs/2005.14165}, 
}

@inproceedings{bai-etal-2024-longbench,
    title = "{L}ong{B}ench: A Bilingual, Multitask Benchmark for Long Context Understanding",
    author = "Bai, Yushi  and
      Lv, Xin  and
      Zhang, Jiajie  and
      Lyu, Hongchang  and
      Tang, Jiankai  and
      Huang, Zhidian  and
      Du, Zhengxiao  and
      Liu, Xiao  and
      Zeng, Aohan  and
      Hou, Lei  and
      Dong, Yuxiao  and
      Tang, Jie  and
      Li, Juanzi",
    editor = "Ku, Lun-Wei  and
      Martins, Andre  and
      Srikumar, Vivek",
    booktitle = "Proceedings of the 62nd Annual Meeting of the Association for Computational Linguistics (ACL)",
    month = aug,
    year = "2024",
    address = "Bangkok, Thailand",
    url = "https://aclanthology.org/2024.acl-long.172/",
    doi = "10.18653/v1/2024.acl-long.172",
    pages = "3119--3137"
}

@misc{openaireasoning,
    title={Learning to reason with {LLM}s},
    author={OpenAI},
    year={2024},
    url={https://openai.com/index/learning-to-reason-with-llms/},
}

@inproceedings{zhang-etal-2025-query-focused,
    title = "Query-Focused Retrieval Heads Improve Long-Context Reasoning and Re-ranking",
    author = "Zhang, Wuwei  and
      Yin, Fangcong  and
      Yen, Howard  and
      Chen, Danqi  and
      Ye, Xi",
    editor = "Christodoulopoulos, Christos  and
      Chakraborty, Tanmoy  and
      Rose, Carolyn  and
      Peng, Violet",
    booktitle = "Proceedings of the 2025 Conference on Empirical Methods in Natural Language Processing (EMNLP)",
    year = "2025",
    url = "https://aclanthology.org/2025.emnlp-main.1214/",
    doi = "10.18653/v1/2025.emnlp-main.1214",
    pages = "23802--23816",
    ISBN = "979-8-89176-332-6"
}

@inproceedings{tang-etal-2024-language,
    title = "Language-Specific Neurons: The Key to Multilingual Capabilities in Large Language Models",
    author = "Tang, Tianyi  and
      Luo, Wenyang  and
      Huang, Haoyang  and
      Zhang, Dongdong  and
      Wang, Xiaolei  and
      Zhao, Xin  and
      Wei, Furu  and
      Wen, Ji-Rong",
    editor = "Ku, Lun-Wei  and
      Martins, Andre  and
      Srikumar, Vivek",
    booktitle = "Proceedings of the 62nd Annual Meeting of the Association for Computational Linguistics (ACL)",
    year = "2024",
    url = "https://aclanthology.org/2024.acl-long.309/",
    doi = "10.18653/v1/2024.acl-long.309",
    pages = "5701--5715"
}

@inproceedings{dai-etal-2022-knowledge,
    title = "Knowledge Neurons in Pretrained Transformers",
    author = "Dai, Damai  and
      Dong, Li  and
      Hao, Yaru  and
      Sui, Zhifang  and
      Chang, Baobao  and
      Wei, Furu",
    editor = "Muresan, Smaranda  and
      Nakov, Preslav  and
      Villavicencio, Aline",
    booktitle = "Proceedings of the 60th Annual Meeting of the Association for Computational Linguistics (ACL)",
    year = "2022",
    url = "https://aclanthology.org/2022.acl-long.581/",
    doi = "10.18653/v1/2022.acl-long.581",
    pages = "8493--8502"
}

@inproceedings{mondal-etal-2025-language,
    title = "Language-Specific Neurons Do Not Facilitate Cross-Lingual Transfer",
    author = "Mondal, Soumen Kumar  and
      Sen, Sayambhu  and
      Singhania, Abhishek  and
      Jyothi, Preethi",
    editor = "Drozd, Aleksandr  and
      Sedoc, Jo{\~a}o  and
      Tafreshi, Shabnam  and
      Akula, Arjun  and
      Shu, Raphael",
    booktitle = "The Sixth Workshop on Insights from Negative Results in NLP",
    year = "2025",
    url = "https://aclanthology.org/2025.insights-1.6/",
    doi = "10.18653/v1/2025.insights-1.6",
    pages = "46--62",
    ISBN = "979-8-89176-240-4"
}

@inproceedings{kojima-etal-2024-multilingual,
    title = "On the Multilingual Ability of Decoder-based Pre-trained Language Models: Finding and Controlling Language-Specific Neurons",
    author = "Kojima, Takeshi  and
      Okimura, Itsuki  and
      Iwasawa, Yusuke  and
      Yanaka, Hitomi  and
      Matsuo, Yutaka",
    editor = "Duh, Kevin  and
      Gomez, Helena  and
      Bethard, Steven",
    booktitle = "Proceedings of the 2024 Conference of the North American Chapter of the Association for Computational Linguistics: Human Language Technologies (NAACL)",
    year = "2024",
    url = "https://aclanthology.org/2024.naacl-long.384/",
    doi = "10.18653/v1/2024.naacl-long.384",
    pages = "6919--6971"
}

@inproceedings{
meng2022locating,
title={Locating and Editing Factual Associations in {GPT}},
author={Kevin Meng and David Bau and Alex J Andonian and Yonatan Belinkov},
booktitle={Advances in Neural Information Processing Systems (NeurIPS)},
editor={Alice H. Oh and Alekh Agarwal and Danielle Belgrave and Kyunghyun Cho},
year={2022},
url={https://openreview.net/forum?id=-h6WAS6eE4}
}

@inproceedings{gu-etal-2024-model,
    title = "Model Editing Harms General Abilities of Large Language Models: Regularization to the Rescue",
    author = "Gu, Jia-Chen  and
      Xu, Hao-Xiang  and
      Ma, Jun-Yu  and
      Lu, Pan  and
      Ling, Zhen-Hua  and
      Chang, Kai-Wei  and
      Peng, Nanyun",
    editor = "Al-Onaizan, Yaser  and
      Bansal, Mohit  and
      Chen, Yun-Nung",
    booktitle = "Proceedings of the 2024 Conference on Empirical Methods in Natural Language Processing (EMNLP)",
    year = "2024",
    url = "https://aclanthology.org/2024.emnlp-main.934/",
    doi = "10.18653/v1/2024.emnlp-main.934",
    pages = "16801--16819"
}

@inproceedings{gao-etal-2025-train,
    title = "How to Train Long-Context Language Models (Effectively)",
    author = "Gao, Tianyu  and
      Wettig, Alexander  and
      Yen, Howard  and
      Chen, Danqi",
    editor = "Che, Wanxiang  and
      Nabende, Joyce  and
      Shutova, Ekaterina  and
      Pilehvar, Mohammad Taher",
    booktitle = "Proceedings of the 63rd Annual Meeting of the Association for Computational Linguistics (ACL)",
    year = "2025",
    url = "https://aclanthology.org/2025.acl-long.366/",
    doi = "10.18653/v1/2025.acl-long.366",
    pages = "7376--7399",
    ISBN = "979-8-89176-251-0"
}

@inproceedings{
rafailov2023direct,
title={Direct Preference Optimization: Your Language Model is Secretly a Reward Model},
author={Rafael Rafailov and Archit Sharma and Eric Mitchell and Christopher D Manning and Stefano Ermon and Chelsea Finn},
booktitle={Thirty-seventh Conference on Neural Information Processing Systems (NeurIPS)},
year={2023},
url={https://openreview.net/forum?id=HPuSIXJaa9}
}

@inproceedings{
zhao2024wildchat,
title={WildChat: 1M Chat{GPT} Interaction Logs in the Wild},
author={Wenting Zhao and Xiang Ren and Jack Hessel and Claire Cardie and Yejin Choi and Yuntian Deng},
booktitle={The Twelfth International Conference on Learning Representations (ICLR)},
year={2024},
url={https://openreview.net/forum?id=Bl8u7ZRlbM}
}

@inproceedings{
cheng2025revisiting,
title={Revisiting Reinforcement Learning for {LLM} Reasoning from A Cross-Domain Perspective},
author={Zhoujun Cheng and Shibo Hao and Tianyang Liu and Fan Zhou and Yutao Xie and Feng Yao and Yuexin Bian and Nilabjo Dey and Yonghao Zhuang and Yuheng Zha and Yi Gu and Kun Zhou and Yuqi Wang and Yuan Li and Richard Fan and Jianshu She and Chengqian Gao and Abulhair Saparov and Taylor W. Killian and Haonan Li and Mikhail Yurochkin and Eric P. Xing and Zhengzhong Liu and Zhiting Hu},
booktitle={The Thirty-ninth Annual Conference on Neural Information Processing Systems Datasets and Benchmarks Track (NeurIPS)},
year={2025},
url={https://openreview.net/forum?id=xUBgfvyip3}
}

@inproceedings{
zheng2024lmsyschatm,
title={{LMSYS}-Chat-1M: A Large-Scale Real-World {LLM} Conversation Dataset},
author={Lianmin Zheng and Wei-Lin Chiang and Ying Sheng and Tianle Li and Siyuan Zhuang and Zhanghao Wu and Yonghao Zhuang and Zhuohan Li and Zi Lin and Eric Xing and Joseph E. Gonzalez and Ion Stoica and Hao Zhang},
booktitle={The Twelfth International Conference on Learning Representations (ICLR)},
year={2024},
url={https://openreview.net/forum?id=BOfDKxfwt0}
}

@misc{grattafiori2024llama3herdmodels,
      title={The {Llama} 3 Herd of Models}, 
      author={Aaron Grattafiori and Abhimanyu Dubey and Abhinav Jauhri and Abhinav Pandey and Abhishek Kadian and Ahmad Al-Dahle and Aiesha Letman and Akhil Mathur and Alan Schelten and Alex Vaughan and Amy Yang and Angela Fan and Anirudh Goyal and Anthony Hartshorn and Aobo Yang and Archi Mitra and Archie Sravankumar and Artem Korenev and Arthur Hinsvark and Arun Rao and Aston Zhang and Aurelien Rodriguez and Austen Gregerson and Ava Spataru and Baptiste Roziere and Bethany Biron and Binh Tang and Bobbie Chern and Charlotte Caucheteux and Chaya Nayak and Chloe Bi and Chris Marra and Chris McConnell and Christian Keller and Christophe Touret and Chunyang Wu and Corinne Wong and Cristian Canton Ferrer and Cyrus Nikolaidis and Damien Allonsius and Daniel Song and Danielle Pintz and Danny Livshits and Danny Wyatt and David Esiobu and Dhruv Choudhary and Dhruv Mahajan and Diego Garcia-Olano and Diego Perino and Dieuwke Hupkes and Egor Lakomkin and Ehab AlBadawy and Elina Lobanova and Emily Dinan and Eric Michael Smith and Filip Radenovic and Francisco Guzmán and Frank Zhang and Gabriel Synnaeve and Gabrielle Lee and Georgia Lewis Anderson and Govind Thattai and Graeme Nail and Gregoire Mialon and Guan Pang and Guillem Cucurell and Hailey Nguyen and Hannah Korevaar and Hu Xu and Hugo Touvron and Iliyan Zarov and Imanol Arrieta Ibarra and Isabel Kloumann and Ishan Misra and Ivan Evtimov and Jack Zhang and Jade Copet and Jaewon Lee and Jan Geffert and Jana Vranes and Jason Park and Jay Mahadeokar and Jeet Shah and Jelmer van der Linde and Jennifer Billock and Jenny Hong and Jenya Lee and Jeremy Fu and Jianfeng Chi and Jianyu Huang and Jiawen Liu and Jie Wang and Jiecao Yu and Joanna Bitton and Joe Spisak and Jongsoo Park and Joseph Rocca and Joshua Johnstun and Joshua Saxe and Junteng Jia and Kalyan Vasuden Alwala and Karthik Prasad and Kartikeya Upasani and Kate Plawiak and Ke Li and Kenneth Heafield and Kevin Stone and Khalid El-Arini and Krithika Iyer and Kshitiz Malik and Kuenley Chiu and Kunal Bhalla and Kushal Lakhotia and Lauren Rantala-Yeary and Laurens van der Maaten and Lawrence Chen and Liang Tan and Liz Jenkins and Louis Martin and Lovish Madaan and Lubo Malo and Lukas Blecher and Lukas Landzaat and Luke de Oliveira and Madeline Muzzi and Mahesh Pasupuleti and Mannat Singh and Manohar Paluri and Marcin Kardas and Maria Tsimpoukelli and Mathew Oldham and Mathieu Rita and Maya Pavlova and Melanie Kambadur and Mike Lewis and Min Si and Mitesh Kumar Singh and Mona Hassan and Naman Goyal and Narjes Torabi and Nikolay Bashlykov and Nikolay Bogoychev and Niladri Chatterji and Ning Zhang and Olivier Duchenne and Onur Çelebi and Patrick Alrassy and Pengchuan Zhang and Pengwei Li and Petar Vasic and Peter Weng and Prajjwal Bhargava and Pratik Dubal and Praveen Krishnan and Punit Singh Koura and Puxin Xu and Qing He and Qingxiao Dong and Ragavan Srinivasan and Raj Ganapathy and Ramon Calderer and Ricardo Silveira Cabral and Robert Stojnic and Roberta Raileanu and Rohan Maheswari and Rohit Girdhar and Rohit Patel and Romain Sauvestre and Ronnie Polidoro and Roshan Sumbaly and Ross Taylor and Ruan Silva and Rui Hou and Rui Wang and Saghar Hosseini and Sahana Chennabasappa and Sanjay Singh and Sean Bell and Seohyun Sonia Kim and Sergey Edunov and Shaoliang Nie and Sharan Narang and Sharath Raparthy and Sheng Shen and Shengye Wan and Shruti Bhosale and Shun Zhang and Simon Vandenhende and Soumya Batra and Spencer Whitman and Sten Sootla and Stephane Collot and Suchin Gururangan and Sydney Borodinsky and Tamar Herman and Tara Fowler and Tarek Sheasha and Thomas Georgiou and Thomas Scialom and Tobias Speckbacher and Todor Mihaylov and Tong Xiao and Ujjwal Karn and Vedanuj Goswami and Vibhor Gupta and Vignesh Ramanathan and Viktor Kerkez and Vincent Gonguet and Virginie Do and Vish Vogeti and Vítor Albiero and Vladan Petrovic and Weiwei Chu and Wenhan Xiong and Wenyin Fu and Whitney Meers and Xavier Martinet and Xiaodong Wang and Xiaofang Wang and Xiaoqing Ellen Tan and Xide Xia and Xinfeng Xie and Xuchao Jia and Xuewei Wang and Yaelle Goldschlag and Yashesh Gaur and Yasmine Babaei and Yi Wen and Yiwen Song and Yuchen Zhang and Yue Li and Yuning Mao and Zacharie Delpierre Coudert and Zheng Yan and Zhengxing Chen and Zoe Papakipos and Aaditya Singh and Aayushi Srivastava and Abha Jain and Adam Kelsey and Adam Shajnfeld and Adithya Gangidi and Adolfo Victoria and Ahuva Goldstand and Ajay Menon and Ajay Sharma and Alex Boesenberg and Alexei Baevski and Allie Feinstein and Amanda Kallet and Amit Sangani and Amos Teo and Anam Yunus and Andrei Lupu and Andres Alvarado and Andrew Caples and Andrew Gu and Andrew Ho and Andrew Poulton and Andrew Ryan and Ankit Ramchandani and Annie Dong and Annie Franco and Anuj Goyal and Aparajita Saraf and Arkabandhu Chowdhury and Ashley Gabriel and Ashwin Bharambe and Assaf Eisenman and Azadeh Yazdan and Beau James and Ben Maurer and Benjamin Leonhardi and Bernie Huang and Beth Loyd and Beto De Paola and Bhargavi Paranjape and Bing Liu and Bo Wu and Boyu Ni and Braden Hancock and Bram Wasti and Brandon Spence and Brani Stojkovic and Brian Gamido and Britt Montalvo and Carl Parker and Carly Burton and Catalina Mejia and Ce Liu and Changhan Wang and Changkyu Kim and Chao Zhou and Chester Hu and Ching-Hsiang Chu and Chris Cai and Chris Tindal and Christoph Feichtenhofer and Cynthia Gao and Damon Civin and Dana Beaty and Daniel Kreymer and Daniel Li and David Adkins and David Xu and Davide Testuggine and Delia David and Devi Parikh and Diana Liskovich and Didem Foss and Dingkang Wang and Duc Le and Dustin Holland and Edward Dowling and Eissa Jamil and Elaine Montgomery and Eleonora Presani and Emily Hahn and Emily Wood and Eric-Tuan Le and Erik Brinkman and Esteban Arcaute and Evan Dunbar and Evan Smothers and Fei Sun and Felix Kreuk and Feng Tian and Filippos Kokkinos and Firat Ozgenel and Francesco Caggioni and Frank Kanayet and Frank Seide and Gabriela Medina Florez and Gabriella Schwarz and Gada Badeer and Georgia Swee and Gil Halpern and Grant Herman and Grigory Sizov and Guangyi and Zhang and Guna Lakshminarayanan and Hakan Inan and Hamid Shojanazeri and Han Zou and Hannah Wang and Hanwen Zha and Haroun Habeeb and Harrison Rudolph and Helen Suk and Henry Aspegren and Hunter Goldman and Hongyuan Zhan and Ibrahim Damlaj and Igor Molybog and Igor Tufanov and Ilias Leontiadis and Irina-Elena Veliche and Itai Gat and Jake Weissman and James Geboski and James Kohli and Janice Lam and Japhet Asher and Jean-Baptiste Gaya and Jeff Marcus and Jeff Tang and Jennifer Chan and Jenny Zhen and Jeremy Reizenstein and Jeremy Teboul and Jessica Zhong and Jian Jin and Jingyi Yang and Joe Cummings and Jon Carvill and Jon Shepard and Jonathan McPhie and Jonathan Torres and Josh Ginsburg and Junjie Wang and Kai Wu and Kam Hou U and Karan Saxena and Kartikay Khandelwal and Katayoun Zand and Kathy Matosich and Kaushik Veeraraghavan and Kelly Michelena and Keqian Li and Kiran Jagadeesh and Kun Huang and Kunal Chawla and Kyle Huang and Lailin Chen and Lakshya Garg and Lavender A and Leandro Silva and Lee Bell and Lei Zhang and Liangpeng Guo and Licheng Yu and Liron Moshkovich and Luca Wehrstedt and Madian Khabsa and Manav Avalani and Manish Bhatt and Martynas Mankus and Matan Hasson and Matthew Lennie and Matthias Reso and Maxim Groshev and Maxim Naumov and Maya Lathi and Meghan Keneally and Miao Liu and Michael L. Seltzer and Michal Valko and Michelle Restrepo and Mihir Patel and Mik Vyatskov and Mikayel Samvelyan and Mike Clark and Mike Macey and Mike Wang and Miquel Jubert Hermoso and Mo Metanat and Mohammad Rastegari and Munish Bansal and Nandhini Santhanam and Natascha Parks and Natasha White and Navyata Bawa and Nayan Singhal and Nick Egebo and Nicolas Usunier and Nikhil Mehta and Nikolay Pavlovich Laptev and Ning Dong and Norman Cheng and Oleg Chernoguz and Olivia Hart and Omkar Salpekar and Ozlem Kalinli and Parkin Kent and Parth Parekh and Paul Saab and Pavan Balaji and Pedro Rittner and Philip Bontrager and Pierre Roux and Piotr Dollar and Polina Zvyagina and Prashant Ratanchandani and Pritish Yuvraj and Qian Liang and Rachad Alao and Rachel Rodriguez and Rafi Ayub and Raghotham Murthy and Raghu Nayani and Rahul Mitra and Rangaprabhu Parthasarathy and Raymond Li and Rebekkah Hogan and Robin Battey and Rocky Wang and Russ Howes and Ruty Rinott and Sachin Mehta and Sachin Siby and Sai Jayesh Bondu and Samyak Datta and Sara Chugh and Sara Hunt and Sargun Dhillon and Sasha Sidorov and Satadru Pan and Saurabh Mahajan and Saurabh Verma and Seiji Yamamoto and Sharadh Ramaswamy and Shaun Lindsay and Shaun Lindsay and Sheng Feng and Shenghao Lin and Shengxin Cindy Zha and Shishir Patil and Shiva Shankar and Shuqiang Zhang and Shuqiang Zhang and Sinong Wang and Sneha Agarwal and Soji Sajuyigbe and Soumith Chintala and Stephanie Max and Stephen Chen and Steve Kehoe and Steve Satterfield and Sudarshan Govindaprasad and Sumit Gupta and Summer Deng and Sungmin Cho and Sunny Virk and Suraj Subramanian and Sy Choudhury and Sydney Goldman and Tal Remez and Tamar Glaser and Tamara Best and Thilo Koehler and Thomas Robinson and Tianhe Li and Tianjun Zhang and Tim Matthews and Timothy Chou and Tzook Shaked and Varun Vontimitta and Victoria Ajayi and Victoria Montanez and Vijai Mohan and Vinay Satish Kumar and Vishal Mangla and Vlad Ionescu and Vlad Poenaru and Vlad Tiberiu Mihailescu and Vladimir Ivanov and Wei Li and Wenchen Wang and Wenwen Jiang and Wes Bouaziz and Will Constable and Xiaocheng Tang and Xiaojian Wu and Xiaolan Wang and Xilun Wu and Xinbo Gao and Yaniv Kleinman and Yanjun Chen and Ye Hu and Ye Jia and Ye Qi and Yenda Li and Yilin Zhang and Ying Zhang and Yossi Adi and Youngjin Nam and Yu and Wang and Yu Zhao and Yuchen Hao and Yundi Qian and Yunlu Li and Yuzi He and Zach Rait and Zachary DeVito and Zef Rosnbrick and Zhaoduo Wen and Zhenyu Yang and Zhiwei Zhao and Zhiyu Ma},
      year={2024},
      eprint={2407.21783},
      archivePrefix={arXiv},
      primaryClass={cs.AI},
      url={https://arxiv.org/abs/2407.21783}, 
}

@misc{yang2025qwen3technicalreport,
      title={Qwen3 Technical Report}, 
      author={An Yang and Anfeng Li and Baosong Yang and Beichen Zhang and Binyuan Hui and Bo Zheng and Bowen Yu and Chang Gao and Chengen Huang and Chenxu Lv and Chujie Zheng and Dayiheng Liu and Fan Zhou and Fei Huang and Feng Hu and Hao Ge and Haoran Wei and Huan Lin and Jialong Tang and Jian Yang and Jianhong Tu and Jianwei Zhang and Jianxin Yang and Jiaxi Yang and Jing Zhou and Jingren Zhou and Junyang Lin and Kai Dang and Keqin Bao and Kexin Yang and Le Yu and Lianghao Deng and Mei Li and Mingfeng Xue and Mingze Li and Pei Zhang and Peng Wang and Qin Zhu and Rui Men and Ruize Gao and Shixuan Liu and Shuang Luo and Tianhao Li and Tianyi Tang and Wenbiao Yin and Xingzhang Ren and Xinyu Wang and Xinyu Zhang and Xuancheng Ren and Yang Fan and Yang Su and Yichang Zhang and Yinger Zhang and Yu Wan and Yuqiong Liu and Zekun Wang and Zeyu Cui and Zhenru Zhang and Zhipeng Zhou and Zihan Qiu},
      year={2025},
      eprint={2505.09388},
      archivePrefix={arXiv},
      primaryClass={cs.CL},
      url={https://arxiv.org/abs/2505.09388}, 
}

@inproceedings{bai-etal-2024-longalign,
    title = "{L}ong{A}lign: A Recipe for Long Context Alignment of Large Language Models",
    author = "Bai, Yushi  and
      Lv, Xin  and
      Zhang, Jiajie  and
      He, Yuze  and
      Qi, Ji  and
      Hou, Lei  and
      Tang, Jie  and
      Dong, Yuxiao  and
      Li, Juanzi",
    editor = "Al-Onaizan, Yaser  and
      Bansal, Mohit  and
      Chen, Yun-Nung",
    booktitle = "Findings of the Association for Computational Linguistics: EMNLP 2024",
    month = nov,
    year = "2024",
    address = "Miami, Florida, USA",
    url = "https://aclanthology.org/2024.findings-emnlp.74/",
    doi = "10.18653/v1/2024.findings-emnlp.74",
    pages = "1376--1395"
}

@inproceedings{wu-etal-2025-longattn,
    title = "{L}ong{A}ttn: Selecting Long-context Training Data via Token-level Attention",
    author = "Wu, Longyun  and
      Zhu, Dawei  and
      Zhao, Guangxiang  and
      Yu, Zhuocheng  and
      Ran, Junfeng  and
      Wong, Xiangyu  and
      Sun, Lin  and
      Li, Sujian",
    editor = "Che, Wanxiang  and
      Nabende, Joyce  and
      Shutova, Ekaterina  and
      Pilehvar, Mohammad Taher",
    booktitle = "Findings of the Association for Computational Linguistics: ACL 2025",
    year = "2025",
    url = "https://aclanthology.org/2025.findings-acl.991/",
    doi = "10.18653/v1/2025.findings-acl.991",
    pages = "19367--19380",
    ISBN = "979-8-89176-256-5"
}

@misc{olmo2025olmo3,
      title={Olmo 3}, 
      author={Team Olmo and Allyson Ettinger and Amanda Bertsch and Bailey Kuehl and David Graham and David Heineman and Dirk Groeneveld and Faeze Brahman and Finbarr Timbers and Hamish Ivison and Jacob Morrison and Jake Poznanski and Kyle Lo and Luca Soldaini and Matt Jordan and Mayee Chen and Michael Noukhovitch and Nathan Lambert and Pete Walsh and Pradeep Dasigi and Robert Berry and Saumya Malik and Saurabh Shah and Scott Geng and Shane Arora and Shashank Gupta and Taira Anderson and Teng Xiao and Tyler Murray and Tyler Romero and Victoria Graf and Akari Asai and Akshita Bhagia and Alexander Wettig and Alisa Liu and Aman Rangapur and Chloe Anastasiades and Costa Huang and Dustin Schwenk and Harsh Trivedi and Ian Magnusson and Jaron Lochner and Jiacheng Liu and Lester James V. Miranda and Maarten Sap and Malia Morgan and Michael Schmitz and Michal Guerquin and Michael Wilson and Regan Huff and Ronan Le Bras and Rui Xin and Rulin Shao and Sam Skjonsberg and Shannon Zejiang Shen and Shuyue Stella Li and Tucker Wilde and Valentina Pyatkin and Will Merrill and Yapei Chang and Yuling Gu and Zhiyuan Zeng and Ashish Sabharwal and Luke Zettlemoyer and Pang Wei Koh and Ali Farhadi and Noah A. Smith and Hannaneh Hajishirzi},
      year={2025},
      eprint={2512.13961},
      archivePrefix={arXiv},
      primaryClass={cs.CL},
      url={https://arxiv.org/abs/2512.13961}, 
}

@misc{gemmateam2025gemma3technicalreport,
      title={Gemma 3 Technical Report}, 
      author={Gemma Team and Aishwarya Kamath and Johan Ferret and Shreya Pathak and Nino Vieillard and Ramona Merhej and Sarah Perrin and Tatiana Matejovicova and Alexandre Ramé and Morgane Rivière and Louis Rouillard and Thomas Mesnard and Geoffrey Cideron and Jean-bastien Grill and Sabela Ramos and Edouard Yvinec and Michelle Casbon and Etienne Pot and Ivo Penchev and Gaël Liu and Francesco Visin and Kathleen Kenealy and Lucas Beyer and Xiaohai Zhai and Anton Tsitsulin and Robert Busa-Fekete and Alex Feng and Noveen Sachdeva and Benjamin Coleman and Yi Gao and Basil Mustafa and Iain Barr and Emilio Parisotto and David Tian and Matan Eyal and Colin Cherry and Jan-Thorsten Peter and Danila Sinopalnikov and Surya Bhupatiraju and Rishabh Agarwal and Mehran Kazemi and Dan Malkin and Ravin Kumar and David Vilar and Idan Brusilovsky and Jiaming Luo and Andreas Steiner and Abe Friesen and Abhanshu Sharma and Abheesht Sharma and Adi Mayrav Gilady and Adrian Goedeckemeyer and Alaa Saade and Alex Feng and Alexander Kolesnikov and Alexei Bendebury and Alvin Abdagic and Amit Vadi and András György and André Susano Pinto and Anil Das and Ankur Bapna and Antoine Miech and Antoine Yang and Antonia Paterson and Ashish Shenoy and Ayan Chakrabarti and Bilal Piot and Bo Wu and Bobak Shahriari and Bryce Petrini and Charlie Chen and Charline Le Lan and Christopher A. Choquette-Choo and CJ Carey and Cormac Brick and Daniel Deutsch and Danielle Eisenbud and Dee Cattle and Derek Cheng and Dimitris Paparas and Divyashree Shivakumar Sreepathihalli and Doug Reid and Dustin Tran and Dustin Zelle and Eric Noland and Erwin Huizenga and Eugene Kharitonov and Frederick Liu and Gagik Amirkhanyan and Glenn Cameron and Hadi Hashemi and Hanna Klimczak-Plucińska and Harman Singh and Harsh Mehta and Harshal Tushar Lehri and Hussein Hazimeh and Ian Ballantyne and Idan Szpektor and Ivan Nardini and Jean Pouget-Abadie and Jetha Chan and Joe Stanton and John Wieting and Jonathan Lai and Jordi Orbay and Joseph Fernandez and Josh Newlan and Ju-yeong Ji and Jyotinder Singh and Kat Black and Kathy Yu and Kevin Hui and Kiran Vodrahalli and Klaus Greff and Linhai Qiu and Marcella Valentine and Marina Coelho and Marvin Ritter and Matt Hoffman and Matthew Watson and Mayank Chaturvedi and Michael Moynihan and Min Ma and Nabila Babar and Natasha Noy and Nathan Byrd and Nick Roy and Nikola Momchev and Nilay Chauhan and Noveen Sachdeva and Oskar Bunyan and Pankil Botarda and Paul Caron and Paul Kishan Rubenstein and Phil Culliton and Philipp Schmid and Pier Giuseppe Sessa and Pingmei Xu and Piotr Stanczyk and Pouya Tafti and Rakesh Shivanna and Renjie Wu and Renke Pan and Reza Rokni and Rob Willoughby and Rohith Vallu and Ryan Mullins and Sammy Jerome and Sara Smoot and Sertan Girgin and Shariq Iqbal and Shashir Reddy and Shruti Sheth and Siim Põder and Sijal Bhatnagar and Sindhu Raghuram Panyam and Sivan Eiger and Susan Zhang and Tianqi Liu and Trevor Yacovone and Tyler Liechty and Uday Kalra and Utku Evci and Vedant Misra and Vincent Roseberry and Vlad Feinberg and Vlad Kolesnikov and Woohyun Han and Woosuk Kwon and Xi Chen and Yinlam Chow and Yuvein Zhu and Zichuan Wei and Zoltan Egyed and Victor Cotruta and Minh Giang and Phoebe Kirk and Anand Rao and Kat Black and Nabila Babar and Jessica Lo and Erica Moreira and Luiz Gustavo Martins and Omar Sanseviero and Lucas Gonzalez and Zach Gleicher and Tris Warkentin and Vahab Mirrokni and Evan Senter and Eli Collins and Joelle Barral and Zoubin Ghahramani and Raia Hadsell and Yossi Matias and D. Sculley and Slav Petrov and Noah Fiedel and Noam Shazeer and Oriol Vinyals and Jeff Dean and Demis Hassabis and Koray Kavukcuoglu and Clement Farabet and Elena Buchatskaya and Jean-Baptiste Alayrac and Rohan Anil and Dmitry and Lepikhin and Sebastian Borgeaud and Olivier Bachem and Armand Joulin and Alek Andreev and Cassidy Hardin and Robert Dadashi and Léonard Hussenot},
      year={2025},
      eprint={2503.19786},
      archivePrefix={arXiv},
      primaryClass={cs.CL},
      url={https://arxiv.org/abs/2503.19786}, 
}

@inproceedings{vllm,
  title={Efficient Memory Management for Large Language Model Serving with PagedAttention},
  author={Woosuk Kwon and Zhuohan Li and Siyuan Zhuang and Ying Sheng and Lianmin Zheng and Cody Hao Yu and Joseph E. Gonzalez and Hao Zhang and Ion Stoica},
  booktitle={Proceedings of the ACM SIGOPS 29th Symposium on Operating Systems Principles},
  year={2023}
}

@inproceedings{wolf-etal-2020-transformers,
    title = "Transformers: State-of-the-Art Natural Language Processing",
    author = "Wolf, Thomas  and
      Debut, Lysandre  and
      Sanh, Victor  and
      Chaumond, Julien  and
      Delangue, Clement  and
      Moi, Anthony  and
      Cistac, Pierric  and
      Rault, Tim  and
      Louf, Remi  and
      Funtowicz, Morgan  and
      Davison, Joe  and
      Shleifer, Sam  and
      von Platen, Patrick  and
      Ma, Clara  and
      Jernite, Yacine  and
      Plu, Julien  and
      Xu, Canwen  and
      Le Scao, Teven  and
      Gugger, Sylvain  and
      Drame, Mariama  and
      Lhoest, Quentin  and
      Rush, Alexander",
    editor = "Liu, Qun  and
      Schlangen, David",
    booktitle = "Proceedings of the 2020 Conference on Empirical Methods in Natural Language Processing: System Demonstrations (EMNLP)",
    year = "2020",
    url = "https://aclanthology.org/2020.emnlp-demos.6/",
    doi = "10.18653/v1/2020.emnlp-demos.6",
    pages = "38--45"
}

@inproceedings{
zheng2023judging,
title={Judging {LLM}-as-a-Judge with {MT}-Bench and Chatbot Arena},
author={Lianmin Zheng and Wei-Lin Chiang and Ying Sheng and Siyuan Zhuang and Zhanghao Wu and Yonghao Zhuang and Zi Lin and Zhuohan Li and Dacheng Li and Eric Xing and Hao Zhang and Joseph E. Gonzalez and Ion Stoica},
booktitle={Thirty-seventh Conference on Neural Information Processing Systems Datasets and Benchmarks Track (NeurIPS)},
year={2023},
url={https://openreview.net/forum?id=uccHPGDlao}
}

@inproceedings{
rein2024gpqa,
title={{GPQA}: A Graduate-Level Google-Proof Q\&A Benchmark},
author={David Rein and Betty Li Hou and Asa Cooper Stickland and Jackson Petty and Richard Yuanzhe Pang and Julien Dirani and Julian Michael and Samuel R. Bowman},
booktitle={First Conference on Language Modeling (COLM)},
year={2024},
url={https://openreview.net/forum?id=Ti67584b98}
}

@article{lightman2023lets,
      title={Let's Verify Step by Step}, 
      author={Lightman, Hunter and Kosaraju, Vineet and Burda, Yura and Edwards, Harri and Baker, Bowen and Lee, Teddy and Leike, Jan and Schulman, John and Sutskever, Ilya and Cobbe, Karl},
      journal={arXiv preprint arXiv:2305.20050},
      year={2023}
}

@misc{chen2021evaluating,
      title={Evaluating Large Language Models Trained on Code}, 
      author={Mark Chen and Jerry Tworek and Heewoo Jun and Qiming Yuan and Henrique Ponde de Oliveira Pinto and Jared Kaplan and Harri Edwards and Yuri Burda and Nicholas Joseph and Greg Brockman and Alex Ray and Raul Puri and Gretchen Krueger and Michael Petrov and Heidy Khlaaf and Girish Sastry and Pamela Mishkin and Brooke Chan and Scott Gray and Nick Ryder and Mikhail Pavlov and Alethea Power and Lukasz Kaiser and Mohammad Bavarian and Clemens Winter and Philippe Tillet and Felipe Petroski Such and Dave Cummings and Matthias Plappert and Fotios Chantzis and Elizabeth Barnes and Ariel Herbert-Voss and William Hebgen Guss and Alex Nichol and Alex Paino and Nikolas Tezak and Jie Tang and Igor Babuschkin and Suchir Balaji and Shantanu Jain and William Saunders and Christopher Hesse and Andrew N. Carr and Jan Leike and Josh Achiam and Vedant Misra and Evan Morikawa and Alec Radford and Matthew Knight and Miles Brundage and Mira Murati and Katie Mayer and Peter Welinder and Bob McGrew and Dario Amodei and Sam McCandlish and Ilya Sutskever and Wojciech Zaremba},
      year={2021},
      eprint={2107.03374},
      archivePrefix={arXiv},
      primaryClass={cs.LG},
      url={https://arxiv.org/abs/2107.03374}, 
}

@inproceedings{
wang2024mmlupro,
title={{MMLU}-Pro: A More Robust and Challenging Multi-Task Language Understanding Benchmark},
author={Yubo Wang and Xueguang Ma and Ge Zhang and Yuansheng Ni and Abhranil Chandra and Shiguang Guo and Weiming Ren and Aaran Arulraj and Xuan He and Ziyan Jiang and Tianle Li and Max Ku and Kai Wang and Alex Zhuang and Rongqi Fan and Xiang Yue and Wenhu Chen},
booktitle={The Thirty-eight Conference on Neural Information Processing Systems Datasets and Benchmarks Track (NeurIPS)},
year={2024},
url={https://openreview.net/forum?id=y10DM6R2r3}
}

@inproceedings{hiraoka-inui-2025-repetition,
    title = "Repetition Neurons: How Do Language Models Produce Repetitions?",
    author = "Hiraoka, Tatsuya  and
      Inui, Kentaro",
    editor = "Chiruzzo, Luis  and
      Ritter, Alan  and
      Wang, Lu",
    booktitle = "Proceedings of the 2025 Conference of the Nations of the Americas Chapter of the Association for Computational Linguistics: Human Language Technologies (NAACL)",
    year = "2025",
    url = "https://aclanthology.org/2025.naacl-short.41/",
    doi = "10.18653/v1/2025.naacl-short.41",
    pages = "483--495",
    ISBN = "979-8-89176-190-2"
}

@misc{snell2024scalingllmtesttimecompute,
      title={Scaling {LLM} Test-Time Compute Optimally can be More Effective than Scaling Model Parameters}, 
      author={Charlie Snell and Jaehoon Lee and Kelvin Xu and Aviral Kumar},
      year={2024},
      eprint={2408.03314},
      archivePrefix={arXiv},
      primaryClass={cs.LG},
      url={https://arxiv.org/abs/2408.03314}, 
}

@inproceedings{
loshchilov2018decoupled,
title={Decoupled Weight Decay Regularization},
author={Ilya Loshchilov and Frank Hutter},
booktitle={International Conference on Learning Representations (ICLR)},
year={2019},
url={https://openreview.net/forum?id=Bkg6RiCqY7},
}

@inproceedings{
hsieh2024ruler,
title={{RULER}: What{\textquoteright}s the Real Context Size of Your Long-Context Language Models?},
author={Cheng-Ping Hsieh and Simeng Sun and Samuel Kriman and Shantanu Acharya and Dima Rekesh and Fei Jia and Boris Ginsburg},
booktitle={First Conference on Language Modeling},
year={2024},
url={https://openreview.net/forum?id=kIoBbc76Sy}
}

@inproceedings{zhang-etal-2025-longreward,
    title = "{L}ong{R}eward: Improving Long-context Large Language Models with {AI} Feedback",
    author = "Zhang, Jiajie  and
      Hou, Zhongni  and
      Lv, Xin  and
      Cao, Shulin  and
      Hou, Zhenyu  and
      Niu, Yilin  and
      Hou, Lei  and
      Dong, Yuxiao  and
      Feng, Ling  and
      Li, Juanzi",
    editor = "Che, Wanxiang  and
      Nabende, Joyce  and
      Shutova, Ekaterina  and
      Pilehvar, Mohammad Taher",
    booktitle = "Proceedings of the 63rd Annual Meeting of the Association for Computational Linguistics (ACL)",
    month = jul,
    year = "2025",
    publisher = "Association for Computational Linguistics",
    url = "https://aclanthology.org/2025.acl-long.187/",
    doi = "10.18653/v1/2025.acl-long.187",
    pages = "3718--3739",
    ISBN = "979-8-89176-251-0"
}

@inproceedings{huang-etal-2025-improving,
    title = "Improving Contextual Faithfulness of Large Language Models via Retrieval Heads-Induced Optimization",
    author = "Huang, Lei  and
      Feng, Xiaocheng  and
      Ma, Weitao  and
      Fan, Yuchun  and
      Feng, Xiachong  and
      Ye, Yangfan  and
      Zhong, Weihong  and
      Gu, Yuxuan  and
      Wang, Baoxin  and
      Wu, Dayong  and
      Hu, Guoping  and
      Qin, Bing",
    editor = "Che, Wanxiang  and
      Nabende, Joyce  and
      Shutova, Ekaterina  and
      Pilehvar, Mohammad Taher",
    booktitle = "Proceedings of the 63rd Annual Meeting of the Association for Computational Linguistics (ACL)",
    month = jul,
    year = "2025",
    url = "https://aclanthology.org/2025.acl-long.826/",
    doi = "10.18653/v1/2025.acl-long.826",
    pages = "16896--16913",
    ISBN = "979-8-89176-251-0"
}

\appendix

\section{Details of Experiment Settings}
\label{appdx:hypara}

\paragraph{Implementation Details.} For Retrieval Head Detection, we use the official implementation from~\citet{wu2025retrieval}.
For Contrastive Response Generation, we deploy models using the vLLM engine~\cite{vllm} for efficient inference.
For Preference Optimization, we use the Transformer Reinforcement Learning (TRL) library\footnote{https://github.com/huggingface/trl}.
For evaluation, we run Llama-3.1 using the default Transformers library~\cite{wolf-etal-2020-transformers} and use the vLLM engine to speed up inference for Qwen3 and Olmo-3, as they produce reasoning content.
For each experiment, we evaluate over a single training run.

\paragraph{Computational Resources.} 
For all models tested in this study, the retrieval head detection and deactivation step finishes in 2 GPU hours (NVIDIA H100).
The contrastive response generation step finishes in 12 and 36 GPU hours (NVIDIA H100) for instruction models (Llama-3.1-8B-Instruct and Olmo-3-7B-Instruct) and reasoning models (Qwen3-8B and Olmo-3-7B-Think), respectively.
The DPO training runs are conducted on either 4$\times$ NVIDIA H100 GPUs or 8$\times$ NVIDIA H200 GPUs, where all runs finish within 24 hours.

\paragraph{Hyper-Parameters.} We train models using the AdamW~\cite{loshchilov2018decoupled} algorithm, with $\beta_1=0.9$ and $\beta_2=0.95$.
For the first 10\% of training steps, we use a linear warmup to gradually increase the learning rate from 0 to 5e-7, then employ a cosine scheduler with a minimal learning rate set to 5e-8.
We also introduce a weight decay of 0.1.
The global batch size is 512.
We only tune the learning rate from a range of \{2.5e-5, 2.5e-6, 5e-7\} using Llama-3.1 and apply the best learning rate on other models.

\paragraph{LLM-As-A-Judge.}
Both HELMET and MT-Bench utilize LLM-As-a-Judge for evaluation.
For HELMET, we follow the original setting and utilize \textit{gpt-4o-2024-05-13} for evaluation.
For MT-Bench, we employ \textit{gpt-4o-2024-08-06} for evaluation.

\paragraph{Threshold $\tau$.}
We tune the threshold $\tau$ from \{0.05, 0.10\} for each model and report the better-performing setting in the main results.
As a pilot experiment, we evaluate models on HELMET without relying on LLM-based judges.
Specifically, we use ROUGE as a proxy for LLM-as-a-Judge to pre-evaluate model performance at a reduced cost.
The results are shown in Table~\ref{tab:pilot}.

\begin{table}
  \centering
  \small
    \begin{tabular}{lcccc}
    \Xhline{3\arrayrulewidth}
    \textbf{Threshold} & \textbf{Llama} & \textbf{Qwen} & \textbf{Olmo(I) } & \textbf{Olmo(T)}\\
    $\tau$ & \textbf{128K} & \textbf{128K} & \textbf{64K} & \textbf{64K} \\
    \Xhline{2\arrayrulewidth}
    --  & 44.88 & 40.62 & 22.98 & 31.89 \\
    \hdashline
    0.05 & 45.92  &  \textbf{41.82} & \textbf{23.56} & \textbf{32.40}   \\
    0.10 & \textbf{46.38} & 40.78 & 23.03 & 31.40 \\
    \Xhline{3\arrayrulewidth}
    \end{tabular}
  \caption{Pilot experiment results on deciding the threshold $\tau$. Llama, Qwen, Olmo(I), Olmo(T) is short for Llama-3.1-8B-Instruct, Qwen3-8B, Olmo-3-7B-Instruct, and Olmo-3-7B-Think, respectively.}
  \label{tab:pilot}
\end{table}

\section{Number of Heads to Mask}
\label{appdx:sweet_spot}

\begin{figure}[t]
    \centering
    \includegraphics[width=1.\linewidth]{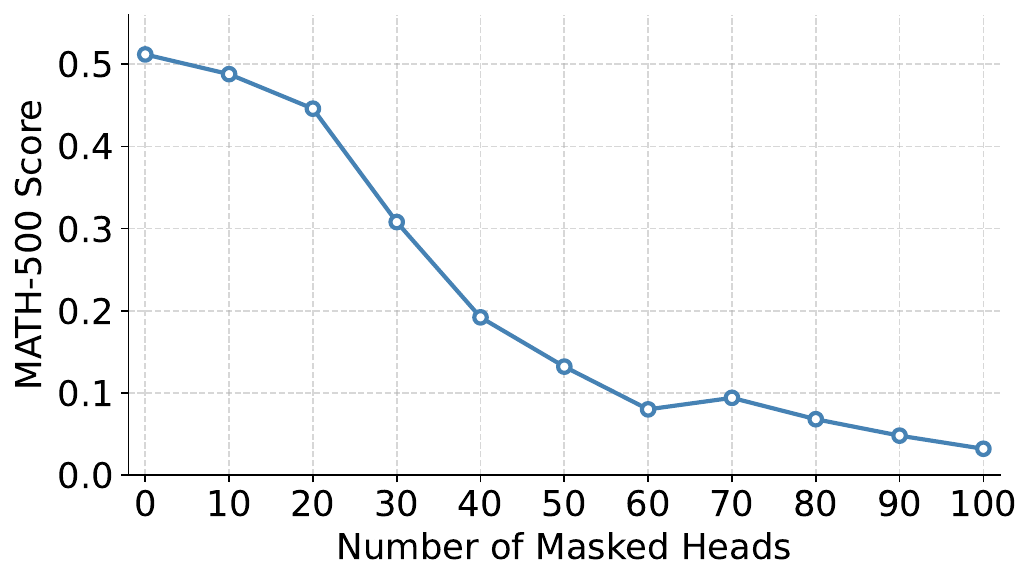}
    \caption{Performance of Llama-3.1-8B-Instruct on MATH-500 when the number of retrieval heads is masked. A sweet spot exists when masking top-30 -- 50 heads, where the model performance degrades but does not collapse.}
    \label{fig:num_heads}
\end{figure}

In order to apply RetMask, it is necessary to set the threshold $\tau$ that determines how many heads are masked.
Masking too few heads results in insufficient contrast between chosen and rejected responses, while masking too many causes rejected responses to degrade into incoherent repetitions.
Although the threshold could in principle be tuned by evaluating trained models, this is computationally expensive.
Instead, as a pilot study, we evaluate the performance of the model with the top-$N$ retrieval-scored heads masked on an external benchmark, and select $\tau$ at the point where performance begins to degrade noticeably.
Figure~\ref{fig:num_heads} shows the results for Llama-3.1-8B-Instruct.

\paragraph{A sweet spot exists when masking the top 30--50 heads.}
As shown in the figure, masking more than 60 heads leads to a sharp performance collapse, with scores dropping below 0.1.
Our chosen threshold $\tau=0.1$ corresponds to masking 40 heads, which falls within a sweet spot that results in enough contrast to yield effective training 
signals, without rendering the ablated model's outputs uninformative.

\section{Qwen3's Task-Wise Performance}
\label{appdx:qwen}

As supplementary material to \S~\ref{exp:main}, we report the task-wise performance of Qwen3-based models evaluated on HELMET in Table~\ref{tab:results_qwen}.
A similar trend to that observed in Table~\ref{tab:results_llama} is evident: RetMask consistently outperforms the baselines on the \textit{Cite} and \textit{Re-rank} tasks.

\begin{table*}[t]
  \centering
  \small
  \begin{tabular}{lc|ccccccc}
  \Xhline{3\arrayrulewidth}
  \textbf{DPO} & \multicolumn{8}{c}{\textbf{Qwen3-8B}}  \\
  \textbf{Strategy} & \textbf{Average} & \textbf{Recall} & \textbf{RAG} & \textbf{Cite }& \textbf{Re-rank} & \textbf{ICL} & \textbf{LongQA} & \textbf{Summ} \\
  \Xhline{2\arrayrulewidth}
  -- & 44.73 & 59.69 & \textbf{53.79} & 12.26 & 15.13 & 82.00 & 47.18 & \textbf{43.06}\\
  \hdashline
  Smaller-Model & 45.51 & 59.56 & 53.54 & 12.42 & 16.86 & \textbf{82.84} & 49.03 & 44.32   \\
  Win-Lose-Pair &  44.49 & 58.63 & 53.33 & 12.59 & 15.17 & 82.16 & 47.39 & 42.18\\
  Non-Retrieval-Mask & 45.48 & 60.25 & 53.17 & 12.53 & 16.08 & 82.28 & \textbf{51.73} & 42.31 \\
  Random-Mask & 45.37 & 60.63 & 53.54 & 13.51 & 16.69 & 82.32 & 47.93 & 39.62  \\
  \hline
  RetMask & \textbf{45.62} & \textbf{60.81} & 53.58 & \textbf{14.74} & \textbf{17.06} & 81.44 & 49.43 & 42.25 \\
  \Xhline{3\arrayrulewidth}
  \end{tabular}
  \caption{Model performance on each task of HELMET when the input sequence length is 128K. The advantage of RetMask is evident on real-world tasks such as generation with citation and passage re-ranking.}
  \label{tab:results_qwen}
\end{table*}

\begin{table}
  \centering
  \small
    \begin{tabular}{lccccc}
    \Xhline{3\arrayrulewidth}
    \textbf{DPO} & \multicolumn{5}{c}{\textbf{Llama-3.1-8B-Instruct}} \\
    \textbf{Strategy} & \textbf{8K} & \textbf{16K} & \textbf{32K} & \textbf{64K} & \textbf{128K} \\
    \Xhline{2\arrayrulewidth}
    -- & 56.03 & 54.14 & 52.42 & 51.65 & 46.40 \\
    \hdashline
    SFT & 53.51 & 50.90 & 49.08 & 44.73 &  37.34 \\
    RetMask & \textbf{58.14} & \textbf{56.92} & \textbf{53.48} & \textbf{53.15} & \textbf{48.68}\\
    \Xhline{3\arrayrulewidth}
    \end{tabular}

  \caption{Llama-3.1-8B-Instruct trained with different strategies, evaluated on HELMET. Models are evaluated using input sequences of 8K, 16K, 32K, and 64K tokens. Training with SFT degrades the performance.}
  \label{tab:sft}
\end{table}

\begin{table}
  \centering
  \small
    \begin{tabular}{lccccc}
    \Xhline{3\arrayrulewidth}
    \textbf{Rejected} & \multicolumn{5}{c}{\textbf{Qwen3-8B}} \\
    \textbf{Samples} & \textbf{8K} & \textbf{16K} & \textbf{32K} & \textbf{64K} & \textbf{128K} \\
    \Xhline{2\arrayrulewidth}
    -- & 50.89 & 47.84 & 47.22 & 42.15 & 40.62 \\
    \hdashline
    Llama-3.1 & 51.85 & \textbf{49.30} & 47.95 & 43.13 & 40.71 \\
    Qwen3 & \textbf{52.40} & 48.88 &  \textbf{48.04} & \textbf{43.39} & \textbf{41.34} \\
    \Xhline{3\arrayrulewidth}
    \end{tabular}

  \caption{Qwen3-8B trained with data synthesized from Llama-3.1-8B-Instruct and Qwen3-8B, evaluated on HELMET. Models are evaluated using input sequences of 8K, 16K, 32K, and 64K tokens. Data synthesized from the target LLM performs better than that synthesized from another LLM in general.}
  \label{tab:syn_diff}
\end{table}

\begin{table*}[t]
  \centering
  \small
  \begin{tabular}{lcccccccc}
  \Xhline{3\arrayrulewidth}
  \textbf{DPO} & \multicolumn{8}{c}{\textbf{Qwen3-8B}}  \\
  \textbf{Strategy} & \textbf{Average} & \textbf{Recall} & \textbf{RAG} & \textbf{Cite} & \textbf{Re-rank} & \textbf{ICL} & \textbf{LongQA} & \textbf{Summ} \\
  \Xhline{2\arrayrulewidth}
  -- & 40.62 & 59.69 & \textbf{53.79} & 12.26 & 15.13 & 82.00 & 43.96 & 17.50\\
  \hdashline
  \multicolumn{8}{l}{\textbf{Trained on LMSYS-Chat-1M}} \\
   RetMask & 41.34 & 61.19 & 52.96 & 12.33 & 16.03 & 82.00 & 46.57 & \textbf{18.30} \\
   \hline
    \multicolumn{8}{l}{\textbf{Trained on Guru-RL-92K}} \\
  Non-Retrieval-Mask & 40.54 & 59.00 & 53.50 & 12.38 & 15.03 & \textbf{82.40} & 44.37 & 17.12  \\
  Random-Mask & 41.00 & 59.38 & 53.54 & 11.78 & \textbf{17.52} & 81.52 & 45.89 & 17.39 \\
  \hline
  RetMask & \textbf{41.59} & 60.38 & 53.58 & \textbf{13.53} & 16.93 & 81.92 & \textbf{47.43} & 17.40  \\
  \Xhline{3\arrayrulewidth}
  \end{tabular}
  \caption{Model performance on each task of HELMET when the input sequence length is  128K when training on Guru-RL-92K. Training with Guru slightly outperforms training with LMSYS-Chat-1M.}
  \label{tab:guru}
\end{table*}

\begin{table*}[t]
  \centering
  \small
    \begin{tabular}{llrrr}
    \Xhline{3\arrayrulewidth}
    & \textbf{Synthesize Model} & \textbf{\# Samples} & \textbf{Avg. Input Length} & \textbf{Avg. Output Length}\\
    \Xhline{2\arrayrulewidth}
    LMSYS-Chat-1M~\cite{zheng2024lmsyschatm} & Llama-3.1 & 294,121 & 63.62 & 494.69 \\
    & Qwen3 & 293,460 & 64.59 & 1642.95\\
    & Olmo-3 (I) & 296,224 & 63.71 &  825.17 \\
    & Olmo-3 (T) & 298,308 & 63.94 & 1816.54 \\
    \hline
    WildChat~\cite{zhao2024wildchat} & Llama-3.1 & 280,184 & 311.68 & 633.01 \\
    \hline
    Guru-RL-92K~\cite{cheng2025revisiting} & Qwen3 (non-reason) & 91,134 & 330.17 & 1965.05\\
    \Xhline{3\arrayrulewidth}
    \end{tabular}
  \caption{Statistics of training data utilized in this work. Olmo-3 (I) and Olmo-3 (T) represents Olmo-3-7B-Instruct and Olmo-3-7B-Think, respectively. The average input/output length is calculated after tokenizing with the corresponding tokenizer.}
  \label{tab:training_stats}
\end{table*}

\section{Supervised Fine-Tuning Baseline}
\label{appdx:sft}

Apart from the DPO baselines included in the paper, we test the effectiveness of Supervised Fine-Tuning (SFT) in pilot experiments.
Specifically, given the preference tuples $\{(x, y_w, y_l)\}$, we train models on $(x, y_w)$ only to focus on the contribution of preferred responses without contrastive signals.
The results are shown in Table~\ref{tab:sft}.

\paragraph{SFT degrades performance while RetMask improves it.}
Table~\ref{tab:sft} shows that SFT on $y_w$ degrades performance below baseline for all input lengths,
while RetMask improves it substantially.
This occurs because training on $y_w$ provides minimal signal: The model learns to reproduce existing behavior without targeted improvement.
In contrast, RetMask succeeds by contrasting $y_w$ with retrieval-degraded $y_l$, thereby creating an optimization objective specifically tailored to retrieval mechanisms.
This validates that RetMask's effectiveness stems from contrastive signals
rather than preferred response quality alone.

\section{Synthesizing Data with Different LLMs}
\label{appdx:syn_diff}

Throughout this paper, we synthesize contrastive training data from the target model itself --- generating both $y_w$ (from the full model $\theta$) and $y_l$ (from the ablated variant $\theta'$) using the same model being trained.
This section examines whether synthesizing data from a more robust model would enhance RetMask's effectiveness.

\paragraph{Settings.} We test cross-model synthesis by training Qwen3-8B on data synthesized from Llama-3.1-8B-Instruct.
This represents a favorable scenario for cross-model synthesis:
(1) Llama-3.1 exhibits stronger baseline long-context capabilities than Qwen3,
and (2) RetMask achieves larger improvements on Llama-3.1 (+2.28) than Qwen3, suggesting higher-quality training signals.
We evaluate using ROUGE scores as a proxy for LLM-as-a-Judge to reduce computational costs.
Results are reported in Table~\ref{tab:syn_diff}.

\paragraph{Self-synthesis outperforms cross-model synthesis.}
Table~\ref{tab:syn_diff} shows that both self-synthesis and cross-model synthesis improve over the baseline, with self-synthesis achieving marginally better results in most settings.
While the performance difference is modest, this pattern suggests that RetMask's training signals are somewhat model-specific:
Masking patterns from one model's retrieval organization may not perfectly align with those of another model, although the transfer is not entirely ineffective.
This indicates that while self-synthesis is preferable for optimal results, cross-model synthesis remains a viable option when computational constraints limit data generation from the target model.

\section{Experiments with RL datasets}
\label{appdx:rl}

We report experimental results obtained using questions from Guru-RL-92K~\cite{cheng2025revisiting} to synthesize responses.
Unlike the datasets used in the prior sections, Guru-RL-92K is specifically collected for reinforcement learning purposes.
It consists of challenging problems across a wide range of domains, including mathematics, coding, science, logic, simulation, and tabular reasoning.

\paragraph{Model.}
We conduct this experiment using Qwen3 and exclude Llama-3.1 and OLMo-3.
Llama-3.1 is not trained for deep reasoning on complex problem-solving tasks and frequently degenerates into repetitive outputs during response generation in this setting.
For OLMo-3, we observe limited effectiveness of RetMask, which we attribute to the model’s internal organization of retrieval-related capabilities.
We therefore exclude both models from this evaluation.

\paragraph{Settings.}
For Qwen3, enabling the reasoning mode often results in very long generations (exceeding 16K tokens) before reaching a final answer, leading to low inference efficiency.
We further observe that masking retrieval heads amplifies this issue, causing the model to generate even longer sequences and to more frequently degenerate into repetitive outputs.
To control for these effects, we conduct experiments with the reasoning mode turned off and compare the results with those obtained in settings in \S~\ref{exp:reasoning} to assess the effectiveness of training on long responses.
In addition to the results using LMSYS-Chat-1M, we include two baselines for Guru-RL-92K: the Non-Retrieval-Mask baseline and the Random-Mask baseline.
The results are reported in Table~\ref{tab:guru}.
Evaluations here also utilize the ROUGE score as a proxy for LLM-as-a-judge.

\paragraph{Training on long outputs slightly outperforms training on short outputs.}
The benefit of training on Guru-RL-92K is most evident on the LongQA task, consistent with the findings in \S~\ref{exp:reasoning}.
These results indicate that synthesizing training data with the reasoning mode enabled can yield additional performance gains.
However, generating responses with reasoning enabled is substantially more computationally expensive, making it less cost-effective than training on standard instruction-tuning datasets.

\section{Statistics of Training Data}
\label{appdx:training_data}

The statistics of training data are shown in Table~\ref{tab:training_stats}.
The number of training samples differs from those reported in the original paper due to two reasons: (1) We filter out samples with personal identifiable information; (2) Some of the samples encountered failure during the process of data synthesis.

\section{Failure Modes of Responses Generated by Retrieval-Head-Ablated Models}
\label{appdx:examples}

In this section, we provide examples that represent the failure modes of responses generated with retrieval-head-ablated models.
In general, these responses exhibit a distinctive failure mode: the model loses track of what it is supposed to maintain—whether that's an entity name, a format constraint, or a reasoning chain, etc.
Other methods' failures tend to be simpler conceptual errors rather than this kind of mid-generation coherence breakdown. 
We provide examples for each category as in Figures~\ref{fig:entity},~\ref{fig:format}, and~\ref{fig:reasoning}.

\begin{itemize}
        \item \textbf{Entity name} (Figure~\ref{fig:entity}): The model with ablated retrieval heads tends to miss entity names. This is a distinctive failure mode we observe only in RetMask.
    \item \textbf{Format Constraint} (Figure~\ref{fig:format}): The model with retrieval heads ablated fails to follow the format constraint. 
    \item \textbf{Reasoning Chain} (Figure~\ref{fig:reasoning}): The model with retrieval heads ablated loses track of its own reasoning chain and generates inconsistent reasoning traces.
\end{itemize}

\begin{figure}[t]
\centering
\small
\begin{tcolorbox}[title=Instruction, colback=gray!10, colframe=gray!50]
Write an introduction for Chengdu Push Bio-technology Co., Ltd, a chemical 
company at No.8 Wuke West Second Road, Wuhou, Chengdu, Sichuan, China.
\end{tcolorbox}
\begin{tcolorbox}[title=Response: Original Model, colback=blue!5, colframe=blue!30]
Chengdu Push Bio-technology Co., Ltd is a prominent chemical company based 
in the vibrant city of Chengdu \textit{[\ldots]}
\end{tcolorbox}
\begin{tcolorbox}[title=Response: Non-Retrieval-Head-Ablated Model, colback=orange!5, colframe=orange!30]
Chengdu Push Bio-technology Co., Ltd is a prominent chemical company situated 
in the vibrant city of Chengdu, Sichuan, China. Headquartered at No.~8 Wuke 
West Second Road, Wuhou District, Chengdu \textit{[\ldots]}
\end{tcolorbox}
\begin{tcolorbox}[title=Response: Retrieval-Head-Ablated Model, colback=red!5, colframe=red!30]
Located in the vibrant city of Chengdu, Sichuan Province, 
\textcolor{red}{Chengdu} is a major chemical hub \textit{[\ldots]} 
The company's manufacturing base is situated at 
\textcolor{red}{No.8, Wuke Road}, Wuhou District \textit{[\ldots]}
\end{tcolorbox}
\caption{Example outputs from three model variants given the same instruction. 
The non-retrieval-head-ablated model correctly reproduces both the company name 
and address. The retrieval-head-ablated model exhibits a distinctive failure 
mode: it corrupts the company name (``Chengdu Push Bio-technology Co., Ltd'' 
$\to$ \textcolor{red}{``Chengdu''}) and the address 
(``No.8 Wuke West Second Road'' $\to$ \textcolor{red}{``No.8, Wuke Road''}), 
a degradation unique to retrieval head ablation.}
\label{fig:entity}
\end{figure}

\begin{figure}[t]
\centering
\small
\begin{tcolorbox}[title=Instruction, colback=gray!10, colframe=gray!50]
Rate the quality of the following story on a scale of 1 to 100. 
Your Response: \texttt{\{"score": integer\}}.\\
\textit{[story omitted]}
\end{tcolorbox}
\begin{tcolorbox}[title=Response: Original Model, colback=blue!5, colframe=blue!30]
\texttt{\{"score": 20\}}\\
This story is extremely bad due to several reasons: (1) Historical 
inaccuracies: The Roman Empire actually fell in 476 AD \textit{[\ldots]}
\end{tcolorbox}
\begin{tcolorbox}[title=Response: Non-Retrieval-Head-Ablated Model, colback=orange!5, colframe=orange!30]
The story seems to be a historical fiction piece, but it appears to be a 
historical inaccuracy. The Roman Empire fell in 476 AD \textit{[\ldots]}\\
I would rate the quality of the story a 20.
\end{tcolorbox}
\begin{tcolorbox}[title=Response: Retrieval-Head-Ablated Model, colback=red!5, colframe=red!30]
\{"Quality": "Poorly Set Up World", \ldots, 
\textcolor{red}{"Final Score": 90}\}\\
\textcolor{red}{"Final Score"}: 90
\end{tcolorbox}
\caption{Example outputs from three model variants given a format-constrained instruction. The original model correctly follows the required format (\texttt{\{"score": integer\}}) and produces a reasonable score. The non-retrieval-head-ablated model produces a reasonable score but ignores the format constraint. The retrieval-head-ablated model both violates the format constraint and assigns an inflated score of 90, demonstrating a distinctive failure mode unique to retrieval head ablation.}
\label{fig:format}
\end{figure}

\begin{figure}[t]
\centering
\small
\begin{tcolorbox}[title=Instruction, colback=gray!10, colframe=gray!50]
Select the correct choice.\\
\textbf{Q}: What is the difference between a RIB and a FIB?\\
A: The FIB is populated based on RIB content.\\
B: The RIB maintains a mirror image of the FIB.\\
C: The RIB is used to make IP source prefix-based switching decisions.\\
D: The FIB is where all IP routing information is stored.
\end{tcolorbox}
\begin{tcolorbox}[title=Original Model, colback=blue!5, colframe=blue!30]
The correct choice is: \textbf{B}. Explanation: RIB (Routing Information Base) is used to store the routing information learned from various routing protocols.\textit{[\ldots]}
\end{tcolorbox}
\begin{tcolorbox}[title=Non-Retrieval-Head-Ablated Model, colback=orange!5, colframe=orange!30]
The correct answer is \textbf{D}. The FIB (Forwarding Information Base) contains routing information for forwarding packets. \textit{[\ldots]}
\end{tcolorbox}
\begin{tcolorbox}[title=Retrieval-Head-Ablated Model, colback=red!5, colframe=red!30]
The correct answer is \textcolor{red}{D} \textit{[\ldots]}
So, the correct answer is \textcolor{red}{not listed in the choices.} 
However, the closest correct answer is \textcolor{red}{B, but with a slight 
modification}: B: The RIB is populated based on the information stored in 
the FIB. \textit{[\ldots]}
\end{tcolorbox}
\caption{Example outputs from three model variants given a multiple-choice 
instruction. Both the original and non-retrieval-head-ablated models produce 
consistent reasoning and commit to a single answer. The retrieval-head-ablated model exhibits inconsistent reasoning: it initially selects \textcolor{red}{D}, then contradicts itself by claiming the correct answer is not among the choices, and finally proposes a modified version of \textcolor{red}{B}.}
\label{fig:reasoning}
\end{figure}

\end{document}